\def\camerareadyversion{1}
\def\arxivversion{1}
\documentclass{article}

\ifdefined\camerareadyversion
\usepackage[final]{corl_2026}
\else
\ifdefined\arxivversion
\usepackage[preprint]{corl_2026} %
\else
\usepackage{corl_2026} %
\fi
\fi
\usepackage{graphicx}
\usepackage{amsmath}
\usepackage{amssymb}
\usepackage{array}
\usepackage{booktabs}
\usepackage{multirow}
\usepackage{placeins}
\usepackage{xcolor}
\makeatletter
\renewcommand{\@conferencelocation}{Austin, Texas, USA}
\makeatother
\ifdefined\arxivversion
\else
\usepackage{xr-hyper}
\IfFileExists{appendix_main.aux}{%
    \externaldocument[][nocite]{appendix_main}%
}{%
    \makeatletter
    \InputIfFileExists{appendix_refs.tex}{}{}%
    \makeatother
}
\fi
\newcommand{\yes}{\(\checkmark\)}
\newcommand{\no}{\(\times\)}
\providecommand{\projectwebsite}{https://dongwon-son.github.io/tam-project-page/}
\newcommand{\projectwebsiteabstract}{%
  \ifdefined\arxivversion
    \unskip\space(\href{\projectwebsite}{website})
  \fi
}

\makeatletter
\newcommand{\beginmaincompact}{%
  \begingroup
  \setlength{\parskip}{3\p@}%
  \setlength{\topsep}{2\p@ \@plus 0.6\p@ \@minus 1\p@}%
  \setlength{\partopsep}{0.2\p@ \@plus 0.2\p@ \@minus 0.2\p@}%
  \setlength{\itemsep}{0.6\p@ \@plus 0.4\p@ \@minus 0.2\p@}%
  \setlength{\parsep}{0.6\p@ \@plus 0.4\p@ \@minus 0.2\p@}%
  \setlength{\textfloatsep}{5\p@ \@plus 1\p@ \@minus 2\p@}%
  \setlength{\floatsep}{4\p@ \@plus 1\p@ \@minus 1.5\p@}%
  \setlength{\intextsep}{5\p@ \@plus 1\p@ \@minus 2\p@}%
  \setlength{\dbltextfloatsep}{5\p@ \@plus 1\p@ \@minus 2\p@}%
  \setlength{\dblfloatsep}{4\p@ \@plus 1\p@ \@minus 1.5\p@}%
  \setlength{\@nipsabovecaptionskip}{2.5\p@}%
  \setlength{\@nipsbelowcaptionskip}{-1\p@}%
  \setlength{\abovecaptionskip}{\@nipsabovecaptionskip}%
  \setlength{\belowcaptionskip}{\@nipsbelowcaptionskip}%
  \setlength{\abovedisplayskip}{5\p@ \@plus 1.5\p@ \@minus 3\p@}%
  \setlength{\belowdisplayskip}{5\p@ \@plus 1.5\p@ \@minus 3\p@}%
  \setlength{\abovedisplayshortskip}{\z@ \@plus 2\p@}%
  \setlength{\belowdisplayshortskip}{3\p@ \@plus 2\p@ \@minus 2\p@}%
  \renewcommand{\section}{%
    \@startsection{section}{1}{\z@}%
                  {-1.2ex \@plus -0.4ex \@minus -0.2ex}%
                  {0.55ex \@plus 0.15ex}%
                  {\large\bf\raggedright}}%
  \renewcommand{\subsection}{%
    \@startsection{subsection}{2}{\z@}%
                  {-1ex \@plus -0.35ex \@minus -0.2ex}%
                  {0.3ex \@plus 0.1ex}%
                  {\normalsize\bf\raggedright}}%
  \renewcommand{\paragraph}{%
    \@startsection{paragraph}{4}{\z@}%
                  {0.55ex \@plus 0.2ex \@minus 0.15ex}%
                  {-1em}%
                  {\normalsize\bf}}%
}
\newcommand{\finishmaincompact}{\endgroup}
\makeatother

\title{TAM: Torque Adaptation Module for Robust Motion Transfer in Manipulation}

\newcommand{\tamrealplainauthors}{Dongwon Son, Florian Shkurti, Jason Lee, Naman Shah, Beomjoon Kim, Dieter Fox}
\newcommand{\tamrealauthorblock}{%
  \begin{tabular}{c}
    Dongwon Son$^{1,*}$
    \quad Florian Shkurti$^{2,3}$
    \quad Jason Lee$^{2,4}$ \\
    Naman Shah$^{2}$
    \quad Beomjoon Kim$^{1,\dagger}$
    \quad Dieter Fox$^{2,4,\dagger}$ \\
    {\normalfont\small $^{1}$KAIST
    \quad $^{2}$Allen Institute for AI
    \quad $^{3}$University of Toronto
    \quad $^{4}$University of Washington} \\
    {\normalfont\small $^{*}$Work done during an internship at the Allen Institute for AI.} \\
    {\normalfont\small $^{\dagger}$Beomjoon Kim and Dieter Fox jointly supervised this work.}
  \end{tabular}%
}

\ifdefined\arxivversion
  \def\showrealauthors{1}
\fi
\ifdefined\showrealauthors
  \author{\tamrealauthorblock}
  \ifdefined\hypersetup
    \hypersetup{pdfauthor={\tamrealplainauthors},pdftitle={TAM: Torque Adaptation Module for Robust Motion Transfer in Manipulation}}
  \fi
\else
  \author{
    Anonymous Author(s)\\
    Affiliation\\
    Address\\
    \texttt{email}
  }
\fi

\begin{document}
\maketitle

\beginmaincompact

\begin{abstract}
    A policy tuned for one robot often behaves differently on another, whether due to the sim-to-real gap, unknown payloads, or the differing dynamics of two instances of the same robot. In contact-rich, dynamic manipulation, even small motion discrepancies can result in failure to track reference motion, since they disrupt the timing and modes of contact. Common remedies, such as domain randomization or system identification, either produce overly conservative task policies~\cite{xie2021dynamics} or require data that must be recollected for each robot or payload~\cite{gaz2019dynamic}.
    We introduce the \textbf{Torque Adaptation Module (TAM)}, a learned module that adapts the torque commands sent to the robot to match the behavior of an ideal robot. TAM operates between the low-level controller that tracks the policy's actions and the robot's torque interface. It includes a history encoder that embeds proprioceptive history into a compact latent state and a torque adaptor that computes residual torque corrections. Because TAM depends only on proprioceptive history and not on policy observations, rewards, or the action space, the same TAM weights can be reused to adapt policies with different action spaces (joint targets, end-effector targets, or direct torques). The policies themselves do not need to be trained with domain randomization of robot parameters. Instead, we offload the need for domain randomization to TAM by training it entirely in randomized simulation, using multi-robot pretraining followed by a robot-specific fine-tuning step that still requires no real-robot data.
    We evaluate TAM zero-shot on a real Franka Panda robot across dynamic manipulation tasks with unknown payloads that include a vision-based box pushing policy (from RL), a flip policy (from BC), and an MPC ball-on-plate balancing controller. Our experiments show that TAM significantly improves zero-shot real-robot execution compared to online system identification and RMA baselines and enables robust dynamic manipulation performance.\projectwebsiteabstract
\end{abstract}


\section{Introduction}

It is well known that discrepancies between simulated and real versions of a manipulator, different payloads, and even slight differences between instances of the same manipulator can cause significant deviations in motion behavior, leading to task failure~\citep{peng2018simtoreal,stone1986arm,farsoni2018realtime}.
This paper proposes a method that enables robust motion transfer in this broader sense: transferring a policy trained or designed for one reference robot to a target robot whose dynamics differ because of the sim-to-real gap, an added payload, or instance-to-instance variation.

Existing techniques usually address this problem either by adapting the model of the manipulator (for example via system identification or neural actuator modelling~\citep{swevers1997optimal,hwangbo2019learning,zhang2025dynamicsprompts}) or by making the policy robust to such discrepancies via domain randomization~\citep{ramos2019bayessim,andrychowicz2020learning,muratore2022robot}.
While system identification methods can learn an accurate model, they require continual model adaptation as the robot instance or payload changes. Domain randomization is not a panacea either, as it can result in overly conservative policies that are invariant to a broad parameter set~\citep{xie2021dynamics}.

Methods such as UPOSI~\cite{yu2017preparing} and RMA~\cite{kumar2021rma,fu2022deepwholebody,liang2024rapid} reduce some of the shortcomings of the approaches above by learning to infer a latent representation of physical properties from observation history, and then conditioning the policy on the inferred dynamics.
This approach is effective when the policy is trained jointly with the latent state and privileged information, however, it suffers from limited reuse: adding adaptation to a new Reinforcement Learning (RL), Behavior Cloning (BC), or Model Predictive Control (MPC) policy requires retraining or distilling the policy weights with the adaptation branch, which is costly or impossible for closed-source or third-party policies.

To overcome these limitations, we propose the \textbf{Torque Adaptation Module (TAM)}, a trajectory-conditioned torque adaptation layer. Our setting assumes an ideal robot model and a target robot. The ideal model is used to train or design a policy, while the target robot may have different physical parameters from the ideal model. At each timestep, the policy outputs an action $a_t$, and the corresponding low-level controller converts that action into the nominal torque $\tau^0_t$ that would be applied to the ideal model. TAM observes the target robot's joint-position, joint-velocity, and applied-torque history, then adds a residual torque to $\tau^0_t$ so the target robot's motion more closely matches the ideal-model motion under the same nominal torque.

\begin{figure*}[t]
    \centering
    \includegraphics[width=\textwidth]{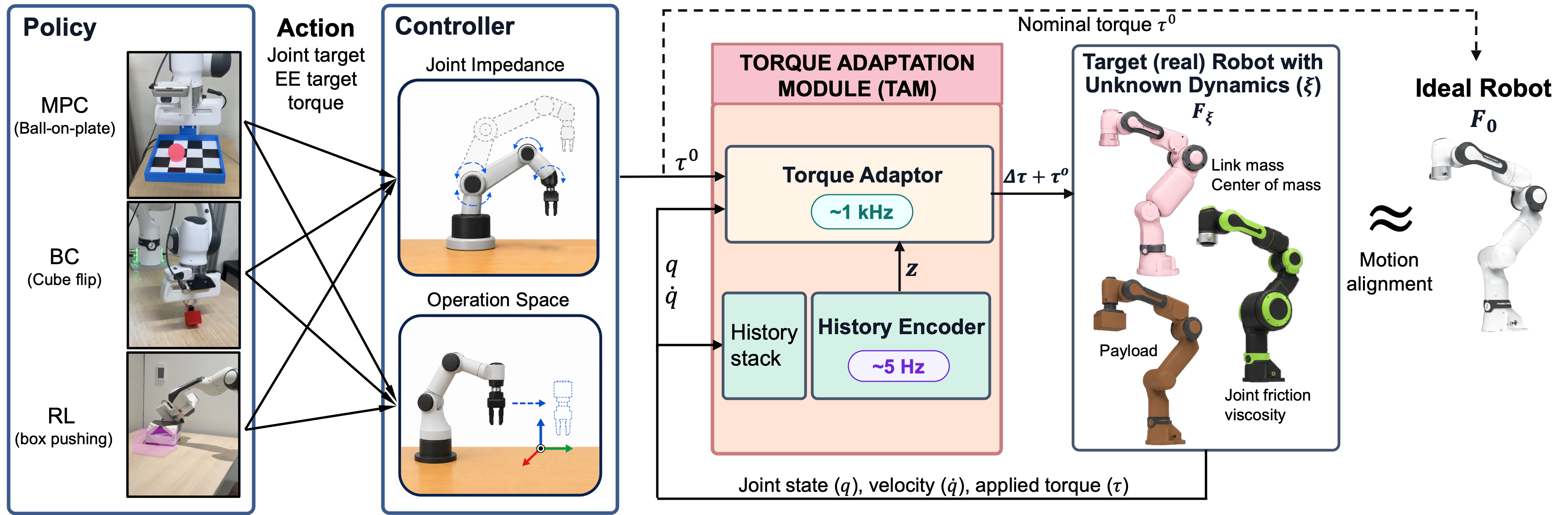}
    \caption{\textbf{TAM deployment interface.}
    Independently authored policies, such as MPC ball balancing, BC cube flipping, or RL box pushing, can be paired with any compatible low-level controller, which converts their actions, including joint targets, end-effector targets, or direct torque, into nominal torque $\tau^0_t$.
    TAM is inserted after this low-level-controller boundary and before the robot plant.
    The history encoder runs asynchronously at about $5\,\mathrm{Hz}$, summarizing the online robot-side history into a latent $z_t$.
    The torque adaptor runs at the $1\,\mathrm{kHz}$ hardware control rate, reusing the most recent $z_t$ to predict the residual torque $\Delta\tau_t$ at every control tick.
    The corrected torque $\tau^0_t+\Delta\tau_t$ is applied to the real robot with unknown joint friction/viscosity, link-mass and center-of-mass mismatch, and payload, so the robot better follows the ideal-model response under the same $\tau^0$ without changing the policy.}
    \label{fig:mam_framework}
    \vspace{-0.3em}
\end{figure*}

\textbf{Contributions.} Our key contribution is to move the adaptation layer below both the policy and the low-level controller, to the torque interface level, as shown in Fig.~\ref{fig:mam_framework}. The first advantage of doing this is that TAM does not require access to policy observations, rewards, task labels, or policy internals. It only requires that the policy action can be converted to nominal torque. Policies can therefore be trained via RL or BC or computed via MPC on the ideal robot without robot-parameter randomization, while TAM handles robot-side mismatch during deployment. The second advantage is that it allows TAM to adapt nominal torques at $1\,\mathrm{kHz}$, which enables fast, dynamic motion.

We evaluate TAM across real-world and simulated robust-transfer settings.
The real-world experiments include tracking with unknown physical parameters and payload, vision-based RL box pushing, a behavior-cloned flip policy, and an MPC ball-on-plate policy, alongside simulated cross-robot tracking across multiple manipulator models.
Across these settings, TAM uses a single shared set of weights to enable real-world transfer of learned RL and BC policies without robot-side domain randomization, and of MPC plans optimized against the ideal model, outperforming online system identification and RMA baselines.

\paragraph{Related Work.}
Appendix~\ref{sec:related} discusses how our work relates to existing literature on sim-to-real transfer, adaptive control, RMA, neural actuator models, policy adaptation, and residual learning.

\section{Problem Definition and Scope}
\label{sec:problem}

We study sim-to-real transfer for \emph{upstream skills} --- policies obtained via reinforcement learning (RL), behavior cloning (BC), or hand-designed via model predictive control (MPC) --- which TAM treats as opaque: it neither accesses nor modifies their internals. These skills are deployed on robots whose dynamics differ from those assumed when the skill was trained.
Let the robot have $N$ actuated joints with joint positions, velocities, and accelerations $q_t,\dot q_t,\ddot q_t\in\mathbb{R}^N$.
A skill is a policy $\pi$ producing an action $a_t=\pi(o_t)\in\mathcal{A}$ from the current observation $o_t$, where the \emph{action space} $\mathcal{A}$ may encode joint targets (with joint impedance), end-effector targets, resolved by operational-space control (OSC), or direct joint torques.
A \emph{low-level controller} $\mathcal{C}:\mathcal{A}\times\mathbb{R}^N\times\mathbb{R}^N\to\mathbb{R}^N$ converts the action together with the current proprioception into a \emph{nominal torque} $\tau^0_t=\mathcal{C}(a_t,q_t,\dot q_t)$.

Let $w_t\in\mathbb{R}^N$ denote an external generalized torque acting on the joints (e.g., a contact wrench $f$ mapped through the manipulator Jacobian transpose $J(q_t)^\top$), and let $F:\mathbb{R}^N\!\times\!\mathbb{R}^N\!\times\!\mathbb{R}^N\!\times\!\mathbb{R}^N\to\mathbb{R}^N$ denote a forward-dynamics map returning the realized acceleration $\ddot q_t=F(q_t,\dot q_t,\tau_t,w_t)$.
The \emph{ideal robot} has known deterministic dynamics $F_0$, used as the reference model while obtaining $\pi$ and $\mathcal{C}$ (e.g., URDF provided by the manufacturer along with rigid-body inertial parameters, no joint friction, etc.). 
The \emph{target robot} has dynamics $F_\xi$, generally stochastic and parameterized by $\xi\in\Xi$ (friction, damping, inertial mismatch, payload, dead zones, actuator bias). On real hardware, $\xi$ is a single fixed but unknown realization; in simulation, we assume that $\xi$ is sampled from a distribution $p_\Xi$ over $\Xi$ during TAM training.
Ideal and target robots share kinematic structure and actuator ordering, so the same $\tau^0_t$ generally produces different accelerations on the two:
$F_\xi(q_t,\dot q_t,\tau^0_t,w_t)\neq F_0(q_t,\dot q_t,\tau^0_t,w_t)$.

To recover the intended outcome on the target robot without retraining $\pi$, we insert a learned module at the \emph{torque interface}, the boundary between $\mathcal{C}$ and the robot.
The TAM module reads $\tau^0_t$ and emits a \emph{residual correction} $\Delta\tau_t\in\mathbb{R}^N$, yielding the \emph{corrected torque}
$\tau^c_t=\operatorname{clip}\!\left(\tau^0_t+\Delta\tau_t,\,\tau_{\min},\,\tau_{\max}\right)$
actually commanded to the robot, where $\tau_{\min},\tau_{\max}\in\mathbb{R}^N$ are the target robot's jointwise actuator limits.
The \emph{applied torque} recorded in the proprioceptive history is then $\tau^a_t=\tau^c_t$.

\textbf{Problem statement.}
\emph{Learn a residual policy $\pi_\Delta$ for torque adaptation, which produces $\Delta\tau_t$ from signals available at deployment such that the target robot driven by $\tau^c_t$ locally matches the ideal robot driven by $\tau^0_t$:}
$F_\xi(q_t,\dot q_t,\tau^c_t,w_t)\approx F_0(q_t,\dot q_t,\tau^0_t,w_t)$.
Note that $w_t$ is \emph{not} observed by $\pi_\Delta$ at deployment; it is treated as known only inside the training-time label generator.

To train $\pi_\Delta$ in simulation, we assume access to inverse torque-command maps $\Phi_0$ and $\Phi_\xi$: given a desired acceleration, $\Phi_\xi(q_t,\dot q_t,\ddot q_t,w_t)$ returns the command torque that produces $\ddot q_t$ under $F_\xi$ and $w_t$, with $\Phi_0$ defined analogously for $F_0$.
These maps are used only to synthesize ground-truth $\Delta\tau$ labels from simulator rollouts; their analytic form is given in Appendix~\ref{app:inverse_command_map} and used in Section~\ref{sec:training_objective}.

\textbf{Scope and limits.}
The setup requires (i) access to the torque interface, (ii) access to $\Phi_\xi$ in simulation for label generation, and (iii) matched kinematic structure and actuator ordering between ideal and target robots.
Clipping enforces command-torque limits but does not by itself guarantee stability or safety: when alignment would require a torque outside $[\tau_{\min},\tau_{\max}]$, $\tau^c_t$ saturates and the residual mismatch cannot be removed, so payloads and contact loads must remain within the actuator budget.

\FloatBarrier

\begin{figure}[!t]
    \centering
    \includegraphics[width=\textwidth]{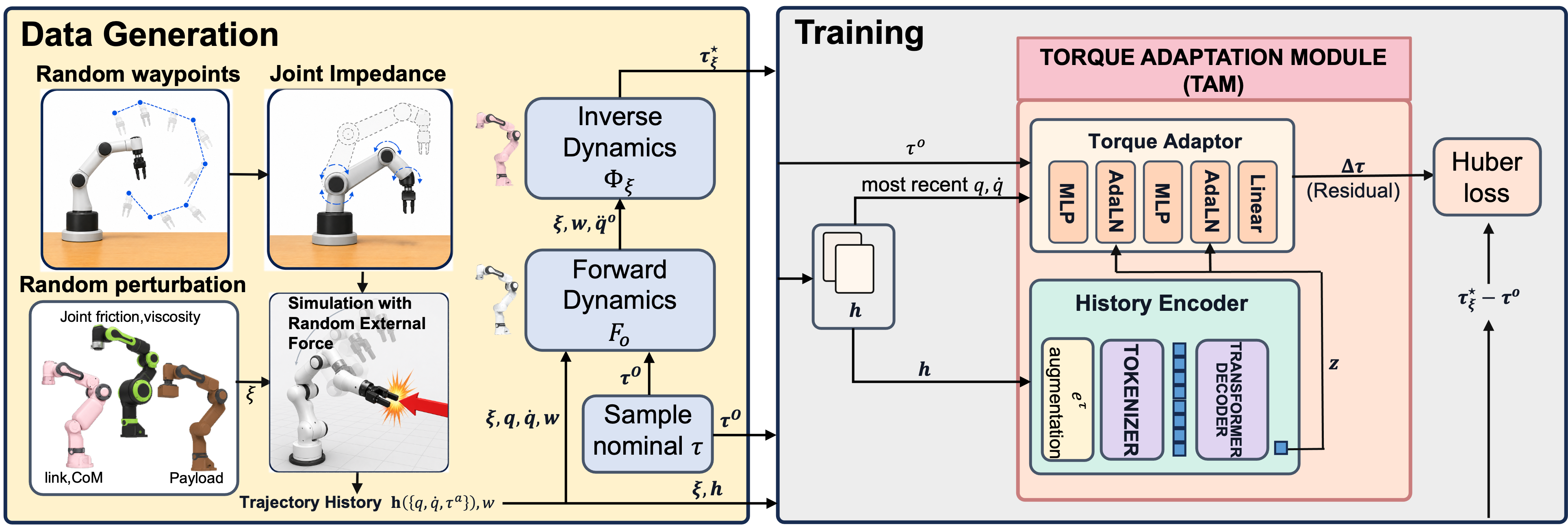}
    \caption{\textbf{TAM data generation and training.}
    Training data are task-agnostic free-space joint-motion rollouts, not downstream task demonstrations.
    For each rollout, we sample perturbation parameters $\xi$ and synthetic randomized joint-space waypoint references designed to cover diverse positions, velocities, torques, and dynamics mismatches.
    A joint impedance low-level controller tracks these waypoints, producing histories $h_t=\{q,\dot q,\tau^a\}$, and we inject smooth external generalized-torque perturbations $w_t$ to mimic the effect of object interaction.
    For each nominal torque $\tau^0_t$, the ideal forward dynamics $F_0$ provides the nominal acceleration, and the perturbed inverse command map $\Phi_\xi$ computes the teacher torque $\tau^\star_{\xi,t}$ needed to realize that ideal response.
    The history encoder maps $h_t$ to a latent $z_t$, the torque adaptor predicts $\Delta\tau_t$, and the Huber loss is applied to the teacher residual $\tau^\star_{\xi,t}-\tau^0_t$.}
    \label{fig:tam_datagen_training}
\end{figure}

\section{Torque Adaptation Module}
\label{sec:method}

\subsection{Data Generation and Training Objective}
\label{sec:training_objective}

Figure~\ref{fig:tam_datagen_training} summarizes the data-generation and training pipeline.
As shown in the figure and caption, task-agnostic joint-motion rollouts produce robot-side histories, while analytic dynamics labels provide the torque target.
The main learning signal is local torque matching: for a recorded state $(q_t,\dot q_t)$ and nominal torque query $\tau^0_t$, the ideal model defines the desired local response through $F_0(q_t,\dot q_t,\tau^0_t,w_t)$, and the teacher corrected torque is the actuator command that makes the perturbed model realize that response:
\begin{equation}
    \tau^\star_\xi(q_t,\dot q_t,\tau^0_t,w_t)
    =
	    \Phi_\xi\!\left(q_t,\dot q_t,F_0(q_t,\dot q_t,\tau^0_t,w_t),w_t\right),
	    \label{eq:gt_torque}
	\end{equation}
	where $\Phi_\xi$ is the inverse torque-command map introduced in Section~\ref{sec:problem} and detailed in Appendix~\ref{app:inverse_command_map}.

Given the sampled nominal torque $\tau^0_t$, let $\hat{\tau}^{c}_{\theta,t}$ denote TAM's corrected-torque prediction.
The target corrected torque is $\tau^{\star}_{\xi,t}=\tau^\star_\xi(q_t,\dot q_t,\tau^0_t,w_t)$.
The loss compares these two corrected torques, $\mathcal{L}_{\tau}=\rho_\delta(\hat{\tau}^{c}_{\theta,t}-\tau^{\star}_{\xi,t})$.
Here $\rho_\delta(\cdot)$ is the elementwise Huber penalty on the joint-torque residual, averaged over joints.
Rollout generation, randomization ranges, nominal-torque query sampling, external-torque handling, the Huber threshold, closed-loop rollout loss, and training-time augmentations are detailed in Appendix~\ref{app:implementation_details}.

Although TAM is trained on randomized motion rollouts rather than downstream task demonstrations, its supervision is not tied to a particular task trajectory.
TAM is trained as a local transition aligner: given $(q_t,\dot q_t)$, robot-side history $h_t$, and $\tau^0_t$, it predicts $\Delta\tau_t$ so the perturbed robot driven by $\tau^c_t$ matches the ideal response $F_0(q_t,\dot q_t,\tau^0_t,w_t)$.
An RL pushing policy, BC flip policy, trajectory-tracking policy, and MPC ball-on-plate policy induce different long-horizon behaviors, but after their actions are converted to nominal torques, they all rely on the same local mapping from joint state and torque to next motion.
Thus TAM requires coverage of relevant state, velocity, torque, and dynamics-mismatch regimes, not coverage of every downstream task trajectory.

\subsection{Encoder--Torque Adaptor Architecture}

The deployment interface in Figure~\ref{fig:mam_framework} highlights the two runtime modules.
The history encoder runs on long-horizon proprioceptive and applied-torque history to infer a compact latent state $z$ that summarizes the current robot-side mismatch.
The torque adaptor runs at the torque interface: it receives $z$, the current state history, and the nominal torque produced by the low-level controller, then outputs the residual torque used to form the corrected torque.
This split lets TAM use long temporal context while keeping the control-rate correction module small enough to attach directly below different policies.

\paragraph{History encoder.}
The history encoder, shown in the TAM network block in Figure~\ref{fig:tam_datagen_training}, summarizes how the target robot has been responding over a long-horizon torque history, which we use as a robustness choice.
For simple tracking motions that keep the joints moving, a shorter window may be enough and would reduce computation.
The longer window helps when the recent few seconds say little about the robot mismatch: a policy may hold almost constant commands, or object interaction may dominate the measured motion.
In these cases, the encoder can still use earlier motion without contact together with the recent history.
At each control step, it first stores a raw rolling buffer of measured position, measured velocity, and applied torque, $\{q_{t-i}, \dot q_{t-i}, \tau^a_{t-i}\}_{i=1}^{K}$.
Here $K$ is the number of controller-rate samples in the history buffer.

The raw buffer already contains the information needed to infer mismatch in principle, but it asks the network to discover inverse-dynamics structure from data alone.
We therefore augment the history with a physics residual feature that answers a more direct question for each past sample: given the measured motion, how much did the applied torque differ from the torque that the ideal model would have needed?
This gives the encoder a past correction-like signal in torque units, which can help infer what residual torque should be applied next.
We define this feature as $e^{\tau}_t=\tau^a_t-\Phi_0(q_t,\dot q_t,\ddot{\hat q}_t,\mathbf{0})$.
Here $\ddot{\hat q}_t$ is the acceleration estimated from local position and velocity traces, and the last argument $\mathbf{0}\in\mathbb{R}^N$ sets the external generalized torque input of the ideal inverse torque-command map to zero.
The encoder input then becomes the augmented rolling history $h_t=\left\{q_{t-i}, \dot q_{t-i}, \tau^a_{t-i}, e^\tau_{t-i}\right\}_{i=1}^{K}$.

We next introduce the neural architecture used to process this long history.
A naive causal Transformer~\citep{vaswani2017attention} over the raw controller-rate sequence would see about $12{,}000$ time samples per joint for the $12\,\mathrm{s}$, $1\,\mathrm{kHz}$ history, making attention memory and computation unnecessarily expensive.
Instead, the encoder follows patch-wise time-series tokenization~\citep{nie2023time}: it splits each joint history into overlapping temporal patches, embeds each patch as a token, and processes the resulting patch-and-joint tokens with a causal Transformer.
This preserves local temporal structure while reducing the sequence length seen by attention.

\paragraph{Torque adaptor.}
The torque adaptor network in Figure~\ref{fig:tam_datagen_training} predicts the residual torque at the control rate from a short local window, the current nominal torque, and the latent state.
It receives $(q,\dot q,\tau^a)$ over the last $L$ control steps, with $L \ll K$, together with the current nominal torque $\tau^0_t$ that should be corrected.
For each joint, separate MLPs embed the local position, velocity, and torque-window features before they are combined with the nominal torque and latent state $z$.
The per-joint latent state $z_{t,j}$ conditions the hidden layers through adaptive layer normalization~\citep{perez2018film,peebles2023dit}, allowing the same torque adaptor weights to express different correction laws under different inferred dynamics.
The torque adaptor then predicts an additive residual torque for each joint,
\begin{equation}
    \Delta\tau_t
    =
    A_\theta(q_{t-L:t},\dot q_{t-L:t},\tau^a_{t-L:t-1},\tau^0_t,z_t).
    \label{eq:adaptor}
\end{equation}
This residual is added to the nominal torque to form the corrected torque $\tau^c_t$, which is applied to the real or perturbed robot.

\subsection{DAgger Fine-Tuning and TAM Residual-Torque Input}
\label{sec:dagger_fusion}

TAM's corrections change the histories encountered during execution, creating a distribution shift from the original training rollouts.
We address this with \emph{DAgger-style fine-tuning}~\citep{ross2011reduction}: we collect simulation rollouts with TAM enabled and further train TAM on the resulting histories using the same analytic teacher-torque objective.
This exposes the encoder to the motion induced by its own corrections, including the smaller tracking errors that arise after adaptation.
Fine-tuning takes place entirely in simulation; the upstream policy remains fixed, and no weight updates or teacher labels are needed on the real robot.

We also evaluate a variant that supplies TAM's past residual-torque outputs $\{\Delta\tau_{t-i}\}_{i=1}^{K}$ as an additional input to the history encoder, alongside the existing history $h_t$.
This additional input records the corrections already attempted, complementing the observed command-to-motion response.
It is distinct from the physics residual $e^\tau$, which is derived from the observed motion and the ideal inverse model.
Both variants retain the same torque-adaptor interface.
Section~\ref{sec:adaptation_latency} compares original TAM, DAgger fine-tuning alone, and DAgger with this additional residual-torque input under abrupt changes in robot dynamics.

\section{Experimental Results}
\label{sec:result}

We evaluate TAM on tasks where the policy is learned for an ideal robot but execution changes when the target robot has different dynamics.
TAM training follows a two-stage simulation pipeline.
First, we pretrain the full TAM module, including the history encoder and torque adaptor, on randomized rollouts pooled from five robot models (Panda, Piper, RBY-1, iiwa14, and Flexiv Rizon4).
This multi-robot pretraining is used to learn a shared torque-correction initialization that can transfer across robot morphologies and serve as the starting point for robot-specific fine-tuning.
Second, when target-robot simulation is available, we fine-tune the same TAM weights on that robot's randomized rollouts.
No policy weights are trained in either stage.
The real-Panda experiments reuse one TAM model fine-tuned in Panda simulation, without DAgger.
Section~\ref{sec:adaptation_latency} compares this model with DAgger fine-tuning alone and DAgger with TAM's residual-torque history as an additional encoder input.
Across those Panda experiments, we change only the policy, its action space or low-level controller, and the target physical condition.
\textbf{We evaluate three hypotheses:} (i) TAM improves motion alignment under unknown physical parameters, low-level-controller changes, and unseen robot models, (ii) the same weights can be attached below policies with different implementations, action spaces, and low-level controllers, and (iii) the history encoder, physics residual feature, and torque adaptor are important design choices.
We present the most task-level real-world results first, then analyze motion alignment and cross-robot simulation behavior; ablations are reported in Appendix~\ref{app:architecture_ablations}.
Table~\ref{tab:experiment_taxonomy} summarizes the experimental structure.

TAM receives only the nominal torque and robot-side history; it does not read policy observations, rewards, task labels, or low-level-controller internals.
No task-specific fine-tuning, policy-specific adaptation, real-world calibration data, or pre-task activation motion is used.
Unless a condition is explicitly labeled as carried-state, adaptation is reset at the start of each trial.
TAM executes the nominal torque directly until the history encoder produces its first latent state, typically after $\approx 0.5\,\mathrm{s}$, and this cold-start interval is included in the reported metric.
Section~\ref{sec:adaptation_latency} evaluates adaptation latency under abrupt mid-episode parameter changes.

\begin{figure}[t]
    \centering
    \includegraphics[width=\linewidth]{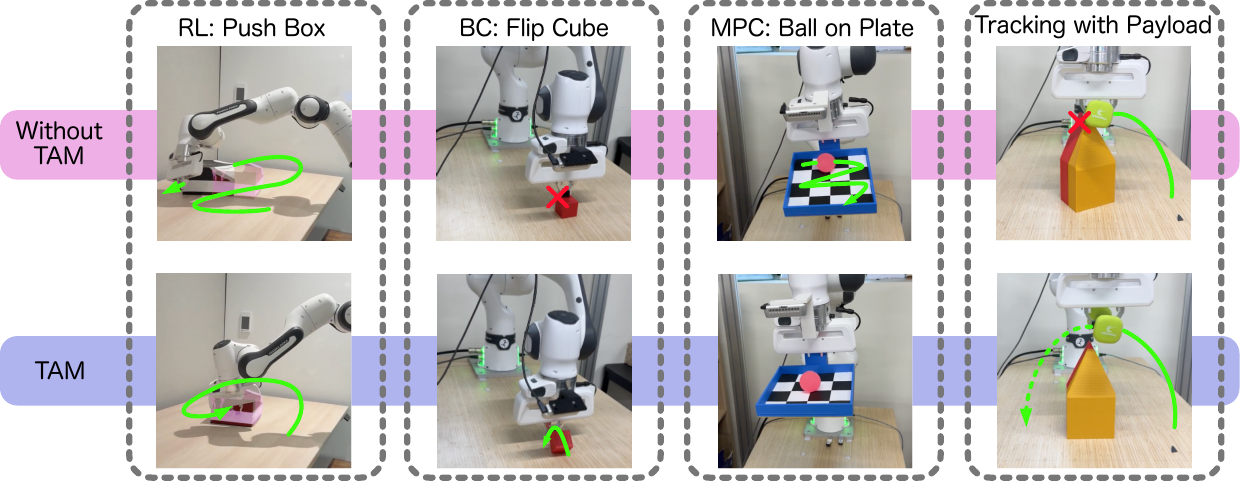}
    \caption{Representative real-world evaluations with direct execution (top, w/o TAM) and TAM-enabled execution (bottom). Columns show the four Panda settings used in the main experiments: RL box pushing, BC cube flipping, an MPC ball-on-plate policy, and tracking with an added payload.}
    \label{fig:realworld_tam_task_overview}
    \vspace{0.6em}
\end{figure}

Figure~\ref{fig:realworld_tam_task_overview} shows the real-world settings.
We compare against Direct, which sends the nominal torque without adaptation, and Online SysID, which estimates explicit physical parameters online with a particle filter and analytically computes the corrected torque.
PhysPred replaces the particle filter with a learned history-to-parameter predictor; for RL pushing, RMA adapts inside the policy rather than at the torque interface.
Baseline details are in Appendix~\ref{app:online_sysid_baseline} and Appendix~\ref{app:vision_box_pushing}.

\subsection{Multiple Policies and Action Spaces}
\label{sec:multiple_upstream_policies}

Table~\ref{tab:realworld_stats} consolidates Direct, Online SysID, RMA, and TAM below the RL, BC, and MPC policies in Table~\ref{tab:experiment_taxonomy}; dashes mark policy-side baselines that are not defined for a task.
Across the three tasks, TAM outperforms all non-ideal baselines.

\begin{table}[t]
    \centering
    {%
    \scriptsize
    \setlength{\tabcolsep}{3.5pt}
    \resizebox{\linewidth}{!}{%
    \begin{tabular}{llcccccc}
        \toprule
        \multirow{2}{*}{Task} & \multirow{2}{*}{Metric} & \multicolumn{1}{c}{Reference} & \multicolumn{5}{c}{Real robot} \\
        \cmidrule(lr){3-3}\cmidrule(lr){4-8}
        & & Ideal sim & Direct & Online SysID & RMA & TAM (w/ RMA) & TAM \\
        \midrule
        Box pushing (depth distill.) & Success rate ($21$ trials) & $84.0\%$ & $47.6\%$ & $57.1\%$ & $52.4\%$ & $66.7\%$ & $\mathbf{76.2\%}$ \\
        Behavior-cloned flip (RGB distill.) & Success rate ($50$ trials) & $73.5\%$ & $50.0\%$ & $34.0\%$ & -- & -- & $\mathbf{72.0\%}$ \\
        Ball-on-plate MPC & Goals / $15\,\mathrm{s}$ ($20$ runs) & $15.9 \pm 2.3$ & $8.8 \pm 2.4$ & $13.1 \pm 2.3$ & -- & -- & $\mathbf{13.6 \pm 2.9}$ \\
        \bottomrule
    \end{tabular}
    }
    }
    \caption{Real-world task statistics for multiple policies. Ideal sim reports the same policy evaluated in its ideal simulation, serving as an idealized upper-bound reference. Trial counts and time limits are given in the metric column; parenthetical labels indicate deployed visual modality. Online SysID is detailed in Appendix~\ref{app:online_sysid_baseline}, and RMA variants are only defined for RL box pushing.
    }
    \label{tab:realworld_stats}
\end{table}

\paragraph{Vision-based pushing policies.}
We adopt the non-prehensile manipulation setup from MuJoCo Playground~\citep{zakka2025mujocoplayground}.
The task goal is to push the box to the commanded planar goal pose within the time limit.
The pushing policy is trained in simulation as a vision policy for the RL pushing task; it consumes a partial point cloud, robot joint position and velocity, and the goal pose.
Appendix~\ref{app:upstream_policy_details} gives the policy-training, deployment, and success-threshold details.
TAM improves over Direct and the strongest non-TAM baseline, Online SysID, by $28.6$ and $19.1$ percentage points, respectively, and also improves the RMA policy from $52.4\%$ to $66.7\%$.
Additional RMA diagnostics are in Appendix~\ref{app:vision_box_pushing}.

\paragraph{Behavior-cloned flip policy.}
The task goal is to execute a contact-rich flip motion that brings the object to the desired final configuration.
The flip policy is an ACT policy~\citep{zhao2023act} trained from demonstrations generated by a scripted demonstrator in ideal-model simulation.
At deployment, the policy receives a gripper-camera image and robot state, outputs joint targets, and a joint-impedance controller converts those targets into the nominal torque corrected by TAM.
Appendix~\ref{app:upstream_policy_details} gives the demonstration-generation and ACT-training details.
This task stresses adaptation because the episode is short, contact timing is sensitive, and the decisive forceful interaction mixes external contact with robot-side mismatch.
As shown in Table~\ref{tab:realworld_stats}, TAM reaches $72\%$ success, close to the $73.5\%$ ideal-simulation baseline.
This improves over direct transfer by $22$ percentage points and over Online SysID by $38$ percentage points, more than doubling its success rate.
Online SysID performs worse than Direct, likely because the short contact-rich policy execution provides poor data for identifying robot parameters: the observed motion reflects both robot mismatch and object interaction.

\paragraph{MPC ball-on-plate.}
The task goal is to repeatedly move the ball to commanded target positions on the plate within a $15\,\mathrm{s}$ time limit.
The ball-on-plate MPC policy uses visual feedback and the ideal model to plan plate motions that move the ball toward target positions with an iLQR planner implemented through MuJoCo MPC~\citep{howell2022predictive}.
This setting tests whether the same robot-side torque adaptor can support a model-based policy rather than a learned policy.
Appendix~\ref{app:upstream_policy_details} gives the estimator, MPC, and goal-counting details.
Direct execution completes $8.8 \pm 2.4$ goals, Online SysID completes $13.1 \pm 2.3$ goals, and TAM completes $13.6 \pm 2.9$ goals, outperforming both baselines.
The ideal-simulation baseline, which represents the policy's performance with no sim-to-real gap, is $15.9 \pm 2.3$ goals per run.

\subsection{Tracking: Motion Alignment Under Unknown Parameters and Low-Level Controller Switches}
\label{sec:tracking_experiments}
\label{sec:cross_controller_tracking}

The tracking experiments isolate motion alignment.
We evaluate real-world end-effector tracking on a Franka Emika Panda after switching from joint impedance to OSC under two physical conditions: the nominal robot and the same robot with an intentionally added $1\,\mathrm{kg}$ end-effector payload.
This test asks whether TAM can make the target robot follow the ideal-model motion more closely when physical parameters or low-level controllers change.
The TAM weights were trained on randomized joint-space waypoint rollouts and did not see any OSC-controlled trajectories during training.

Each trial is split into two segments and uses a sampled source trajectory: for each joint, the sine amplitude and period are randomized to generate a $16\,\mathrm{s}$ joint-space reference tracked by joint impedance control.
At $t=16\,\mathrm{s}$, the source controller is replaced by OSC, which tracks an end-effector pose trajectory for the remaining $8\,\mathrm{s}$ of the trial.
Table~\ref{tab:real_ji_to_osc} evaluates tracking error during this $8\,\mathrm{s}$ OSC segment.
At the OSC switch, \emph{Reset} clears adaptation state, while \emph{Carried} preserves state inferred during the joint-impedance source segment.
Appendix~\ref{app:source_to_osc_protocol} gives the full protocol, additional metrics, and representative carried-state payload trajectories in Figure~\ref{fig:real_panda_payload_xyz_carried}.

\begin{table}[t]
    \centering
    {%
    \scriptsize
    \begingroup
    \setlength{\tabcolsep}{3pt}
    \renewcommand{\arraystretch}{0.9}
    \begin{tabular}{@{}lcccc@{}}
        \toprule
        \multirow{2}{*}{Method} & \multicolumn{2}{c}{Real Panda (hidden params)} & \multicolumn{2}{c}{Real Panda + 1 kg payload} \\
        \cmidrule(lr){2-3}\cmidrule(l){4-5}
        & Reset & Carried & Reset & Carried \\
        \midrule
        Direct & \multicolumn{2}{c}{$4.09 \pm 0.39$} & \multicolumn{2}{c}{$6.04 \pm 0.51$} \\
        Online SysID & $2.67 \pm 0.63$ & $2.75 \pm 0.58$ & $4.69 \pm 0.42$ & $4.65 \pm 0.53$ \\
        PhysPred & $1.74 \pm 0.50$ & $1.64 \pm 0.56$ & $4.15 \pm 0.41$ & $4.39 \pm 0.35$ \\
        TAM (ours) & $1.51 \pm 0.30$ & $\mathbf{1.04 \pm 0.24}$ & $2.35 \pm 0.39$ & $\mathbf{1.29 \pm 0.35}$ \\
        \bottomrule
    \end{tabular}
    \endgroup
    }
    \vspace{0.4em}
    \caption{End-effector target-position tracking under two real-world physical conditions.
    All entries are target-position RMSE against the ideal-model reference in centimeters.
    ``Real Panda (hidden params)'' denotes the physical Franka Panda whose true physical parameters are not provided to the adaptation methods, while ``Real Panda + 1 kg payload'' intentionally adds an unknown $1\,\mathrm{kg}$ end-effector payload.
    For adaptive methods, Reset and Carried denote whether the online adaptation state is cleared or preserved at the start of the OSC segment; Direct has no online state and spans both columns.
    Values are mean $\pm$ standard deviation over $10$ real-world trials per condition. Lower is better.}
    \label{tab:real_ji_to_osc}
\end{table}

TAM gives the lowest tracking error in all four settings.
Carried TAM reduces direct OSC error from $4.09$ to $1.04\,\mathrm{cm}$ on the Panda instance and from $6.04$ to $1.29\,\mathrm{cm}$ with the payload, while remaining below the explicit-parameter baselines.
The reset and carried rows show that TAM can adapt from the target segment alone, but history collected under joint impedance further improves the later OSC segment.

\FloatBarrier

\subsection{Simulation Cross-Robot Generalization}
\label{sec:cross_robot_generalization}

\begin{table}[!t]
    \centering
    {%
    \scriptsize
    \begingroup
    \setlength{\tabcolsep}{3pt}
    \renewcommand{\arraystretch}{0.85}
    \resizebox{0.8\linewidth}{!}{%
    \begin{tabular}{lcccccc}
        \toprule
        Robot & DoF & Split & Direct & Online SysID & TAM & TAM sim-ft. \\
        \midrule
        Panda & $7$ & Train & $4.16 \pm 0.99$ & $1.43 \pm 0.53$ & $\underline{0.59 \pm 0.19}$ & $\mathbf{0.43 \pm 0.17}$ \\
        Piper & $6$ & Train & $4.88 \pm 0.96$ & $0.83 \pm 0.18$ & $\underline{0.52 \pm 0.17}$ & $\mathbf{0.42 \pm 0.15}$ \\
        iiwa14 & $7$ & Train & $3.69 \pm 0.82$ & $0.99 \pm 0.35$ & $\underline{0.56 \pm 0.16}$ & $\mathbf{0.47 \pm 0.16}$ \\
        Flexiv Rizon4 & $7$ & Train & $4.06 \pm 1.20$ & $1.33 \pm 0.51$ & $\underline{0.84 \pm 0.33}$ & $\mathbf{0.68 \pm 0.27}$ \\
        RBY-1 & $7$ & Train & $3.80 \pm 0.81$ & $1.39 \pm 0.83$ & $\underline{0.84 \pm 0.16}$ & $\mathbf{0.66 \pm 0.17}$ \\
        Google Robot & $7$ & Test & $4.69 \pm 1.13$ & $1.65 \pm 0.37$ & $\underline{1.05 \pm 0.40}$ & $\mathbf{0.93 \pm 0.32}$ \\
        Unitree Z1 & $6$ & Test & $6.11 \pm 1.27$ & $1.05 \pm 0.23$ & $\underline{0.93 \pm 0.19}$ & $\mathbf{0.72 \pm 0.24}$ \\
        \midrule
        Mean over robots & -- & -- & $4.48$ & $1.24$ & $\underline{0.76}$ & $\mathbf{0.62}$ \\
        \bottomrule
    \end{tabular}}
    \endgroup
    }
    \vspace{0.4em}
    \caption{Cross-robot tracking transfer in simulation. Values are joint-position RMSE in degrees over $100$ randomized trials; lower is better. Split marks train/test robots, and ``TAM sim-ft.'' is robot-wise simulation fine-tuning from the multi-robot pretrained weights. Bold/underline mark best/second-best.}
    \label{tab:cross_robot_tracking}
\end{table}

We next ask whether TAM can reduce tracking error on robot models beyond the single robot used for the main real-world study.
We use the same randomized tracking evaluation structure across robots, replacing the robot morphology while keeping the low-level-controller and evaluation protocol fixed.
For each test robot, we instantiate its MuJoCo XML asset from MuJoCo Menagerie~\citep{zakka2022mujocomenagerie} and evaluate the multi-robot pretrained weights and robot-wise target-robot simulation fine-tuning.
Appendix~\ref{app:cross_robot_tracking_protocol} details the robot instantiation, DoF handling, and cross-robot $e^\tau$ ablation.
Table~\ref{tab:cross_robot_tracking} reports simulated joint-position tracking RMSE in degrees.

The multi-robot TAM weights reduce tracking error on every reported robot, including the test-only robots, showing that torque-level correction can transfer across simulated robot models.
Simulation fine-tuning gives the lowest mean error across robots ($0.62^\circ$), improving over multi-robot training ($0.76^\circ$) and Online SysID ($1.24^\circ$), while the test-only rows still show useful cross-robot structure without target-robot training.
We therefore interpret multi-robot transfer as promising evidence of reusable correction structure, and simulation fine-tuning as the preferred practical path for obtaining the best target-robot performance.
Appendix~\ref{app:architecture_ablations} reports simulation ablations that isolate the history encoder, physics residual feature, and torque adaptor design choices.

\subsection{Adaptation Latency Under Abrupt Parameter Changes}
\label{sec:adaptation_latency}

\begin{figure}[t]
  \centering
  \includegraphics[width=0.72\linewidth]{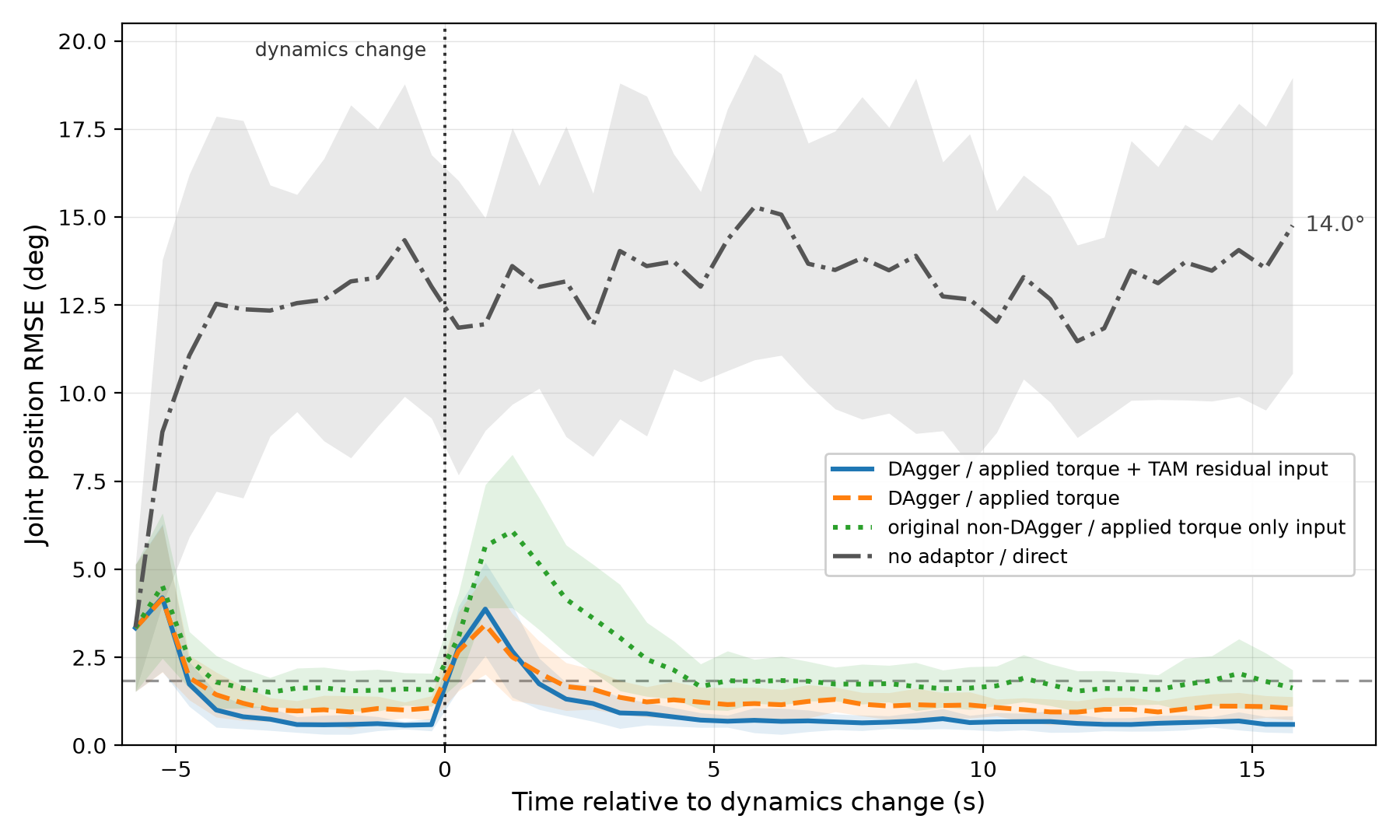}
  \caption{Adaptation after an abrupt dynamics change at $t=0$ over $20$ simulated trajectories. Shading shows variation across trajectories; the dashed line marks original TAM's steady-state RMSE. ``DAgger'' uses simulation fine-tuning on TAM-corrected histories; ``+ TAM residual input'' additionally supplies TAM's past residual-torque outputs to the history encoder.}
  \label{fig:adaptation_latency}
\end{figure}

To evaluate how quickly TAM adapts to changing dynamics, we abruptly switch the simulated robot's physical parameters at $t=0$ in $20$ random trajectories and track joint-position RMSE.
We compare original TAM with DAgger fine-tuning alone and DAgger with TAM's residual-torque history as an additional encoder input (Section~\ref{sec:dagger_fusion}) under the same protocol.
Both variants recover to the original TAM's steady-state error level in $1.7$ s, versus $4.6$ s for original TAM (Figure~\ref{fig:adaptation_latency}).
Their steady-state RMSEs are $1.10^\circ$ and $0.64^\circ$, respectively, compared with $1.83^\circ$ for original TAM and approximately $14^\circ$ for Direct.
Thus fine-tuning on TAM-corrected histories accelerates recovery, while adding TAM's residual-torque history to the encoder input further improves steady tracking accuracy.
Recovery includes trajectory catch-up after the dynamics switch; without an abrupt change, error falls by $90\%$ after the $0.5$ s warm start (Appendix~\ref{app:adaptation_latency}).
All other reported experiments use original TAM.

\section{Limitations}
\label{sec:limitations}

TAM has several scope limitations.
It assumes access to a torque interface, a usable ideal inverse torque-command map, and matched kinematic structure and actuator ordering between the ideal and target robot.
Robots with only position-level APIs or substantially different morphology may require a command-inversion layer or a different torque adaptor design.
The training target is also local: TAM learns the torque needed for one-step motion alignment, not a formal guarantee of long-horizon stability, constraint satisfaction, or task success.
All corrections are clipped to the robot's jointwise torque budget, so payloads or disturbances that require torque beyond hardware authority remain outside scope.
Contact-rich manipulation remains a limitation: randomized external torques bias TAM to preserve external effects, but deployment contacts, object dynamics, or force patterns outside this training distribution may still be confused with robot-side mismatch.

\section{Conclusion}
\label{sec:conclusion}

We presented \textbf{TAM}, a trajectory-conditioned adaptation layer that operates at the torque interface below a policy authored with respect to an ideal robot model: it compresses long-horizon proprioceptive and applied torque history into a compact latent state and adds a residual torque, so the perturbed plant's local motion response matches the ideal robot's response.
With this design, a single shared set of TAM weights enables more robust sim-to-real transfer of learned policies that were not themselves trained with domain randomization, such as a vision-based RL box-pushing policy, a behavior-cloned flip policy, and an MPC ball-on-plate controller.
Our cross-robot tracking study further shows that a single set of TAM weights generalizes across distinct manipulator embodiments, including robots never seen during training, and accuracy can be improved further by fine-tuning on simulated trajectories, requiring no real-world data collection. Overall, we have shown that TAM enables robust, dynamic manipulation.

\FloatBarrier
\clearpage
\acknowledgments{%
This work was partly supported by Institute for Information \& communications Technology Planning \& Evaluation (IITP) grant funded by the Korea government (MSIT) (RS-2019-II190075, Artificial Intelligence Graduate School Program(KAIST)).
This work was also supported by IITP grants funded by the Korea government (MSIT) (RS-2022-II220311, Development of Goal-Oriented Reinforcement Learning Techniques for Contact-Rich Robotic Manipulation of Everyday Objects; RS-2022-II220612, Geometric and Physical Commonsense Reasoning based Behavior Intelligence for Embodied AI; and RS-2024-00457882, National AI Research Hub Project).
This work was supported by the National Research Foundation of Korea (NRF) grant funded by the Korea government (MSIT) (RS-2024-00359085, Foundation model for learning-based humanoid robot that can understand and achieve language commands in unstructured human environments).
This work was supported by the Institute of Information \& Communications Technology Planning \& Evaluation (IITP) with a grant funded by the Ministry of Science and ICT (MSIT) of the Republic of Korea in connection with the Global AI Frontier Lab International Collaborative Research. (No. RS-2024-00469482 \& RS-2024-00509279)
This work was supported by the InnoCORE program of the Ministry of Science and ICT (N10260099).
Additional support was provided by the Ministry of Trade, Industry and Energy (MOTIE) and the Korea Planning \& Evaluation Institute of Industrial Technology (KEIT) under the project \textquotedblleft{}Dexterous and Agile Humanoid Robots for Industrial Applications (DARIUS)\textquotedblright{} (RS-2024-00427719).
}

\ifdefined\arxivversion
\FloatBarrier
\bibliography{ref}  %
\clearpage
\appendix
\renewcommand{\theHsection}{appendix.\Alph{section}}
\renewcommand{\theHsubsection}{appendix.\Alph{section}.\arabic{subsection}}
\makeatletter
\@addtoreset{figure}{section}
\@addtoreset{table}{section}
\@addtoreset{equation}{section}
\makeatother
\renewcommand{\thefigure}{\Alph{section}.\arabic{figure}}
\renewcommand{\theHfigure}{appendix.\Alph{section}.\arabic{figure}}
\renewcommand{\thetable}{\Alph{section}.\arabic{table}}
\renewcommand{\theHtable}{appendix.\Alph{section}.\arabic{table}}
\renewcommand{\theequation}{\Alph{section}.\arabic{equation}}
\renewcommand{\theHequation}{appendix.\Alph{section}.\arabic{equation}}
\newtheorem{tamproposition}{Proposition}[section]

\section{Related Work}
\label{sec:related}
\label{app:related_positioning}

\subsection{Simulator and Actuator Adaptation}
A large body of sim-to-real work addresses the reality gap by constructing a simulator tailored to the target setup, and then training a policy or planning actions in that refined simulator for zero-shot deployment.
Classical robot-identification methods fit dynamic parameters from measured executions or designed excitation trajectories~\citep{swevers1997optimal,gaz2019dynamic}.
Simulator-adaptation methods use these estimates, or a distribution over them, to refine the simulator before downstream policy learning or planning~\citep{ramos2019bayessim}.
Active System Identification (ASID) is a recent manipulation example: it actively collects real-world interaction data, identifies task-relevant physical parameters such as articulation and mass, updates the simulator, and then trains or plans in the refined model~\citep{memmel2024asid}.

Such approaches can be effective, but they depend on a chosen simulator parameterization: the designer must decide which inertial, frictional, contact, or actuator terms should be identified.
This becomes difficult when the dominant mismatch is not well described by a small set of hand-labeled parameters.
In particular, actuator-model mismatch may involve nonlinearities, saturation, delay, or hidden internal states that are not directly observable.
A distinct but related family, \emph{neural actuator models}, addresses this limitation by replacing the analytical actuator equations with a learned model fit from real robot data, and then using the corrected simulator for policy training~\citep{hwangbo2019learning,fey2025uan,liu2025dexndm}.
However, these methods typically require robot-specific interaction data and may need refitting when the new hardware, payload, or actuator condition falls outside the collected data distribution.
More recent in-context system-identification methods, such as Dynamics as Prompts~\citep{zhang2025dynamicsprompts}, reduce this burden by using interaction history to update simulator parameters online without gradient updates.
This is promising because one trained model can adapt across dynamics settings or target environments covered by its parameterization and training distribution.
However, inferring explicit physical parameters from finite history remains ill posed: the observed motion may excite only part of the dynamics, so multiple parameter settings can explain the same transitions and imply the same net torque error~\citep{ljung1999system,swevers1997optimal}.

TAM takes the other route: its history encoder predicts a latent state instead of explicit physical parameters, and the torque adaptor decodes that latent state into the corrected torque command itself.
The latent state is therefore optimized for the information needed to reduce motion mismatch between the ideal and real robot.
This avoids the need to supervise the encoder on explicit physical parameters, while retaining online per-instance adaptation within a single trained module.

\subsection{Policy-Side Adaptation}
Robust policy training with domain or dynamics randomization trains each policy over a distribution of simulated dynamics so the policy itself can tolerate model error~\citep{ramos2019bayessim,andrychowicz2020learning,muratore2022robot}.
This improves robustness, but it places the burden on every task policy: broad randomization can make the learned behavior more conservative or harder to optimize, especially in contact-rich manipulation where the policy must solve the task across many dynamics at once.
Policy-side online adaptation methods address this by inferring an adaptation signal from recent observations, such as body mass, friction, payload, or other dynamics-related context, and conditioning the policy on that estimate.
UPOSI and RMA are representative examples for online system identification and rapid adaptation~\citep{yu2017preparing,kumar2021rma,fu2022deepwholebody}.
This idea has been extended to dexterous manipulation, manipulator-arm policies, long-context locomotion, and disturbance-aware motion tracking~\citep{qi2022inhand,liang2024rapid,liu2025locoformer,zhang2025any2track}.
Policy-side adaptation is highly effective when the policy and task distribution are trained jointly with the adaptation state.
For standard UPOSI/RMA-style instantiations, however, the adaptation embedding is learned as part of that policy's observation/action space and training distribution.
Reusing the embedding below an independently authored policy with a different implementation type or action space is therefore not plug-and-play; a learned policy usually has to be retrained or distilled with its own adaptation input unless it was trained as a broad multi-task policy, and the embedding does not directly apply to a model-based policy.
TAM moves adaptation to the torque interface: policies can be trained or authored for the ideal robot, while TAM uses proprioceptive and applied-torque history to make the target robot execute the resulting nominal torque more like that ideal model.
Policy-side adaptation and TAM therefore act at different levels: policy-side methods adapt the policy action, while TAM adapts the robot's execution of the resulting command.

\FloatBarrier

\FloatBarrier

\subsection{Adaptive Control and Online Identification}
Classical adaptive robot control estimates uncertain dynamics online and can provide tracking guarantees when paired with a structured feedback law~\citep{slotine1987adaptive}.
Recent work also demonstrates that adaptive control can be combined with learned policies; for example, RL2AC integrates an adaptive controller with a reinforcement-learning locomotion policy to compensate changing execution conditions~\citep{lyu2024rl2ac}.
These approaches are attractive when the feedback law is part of the system being designed, but they usually assume a compatible control structure or explicit estimator.
TAM targets a different integration setting: the policy and low-level controller are already fixed, and adaptation is inserted only after their command has been converted to nominal torque.
The relevant advantage is modularity across independently authored controllers and policies: TAM does not require deriving a new adaptive control law, exposing policy observations or rewards, or retraining the policy for each RL, behavior-cloned, MPC, joint-impedance, or OSC policy/controller setup.

Filtering-based real-time identification provides a more modular estimation route.
Constrained Extended Kalman filtering (EKF) has been used to estimate manipulator and load dynamic parameters online, and Kalman-filter pipelines have been used for real-time robot payload identification~\citep{joukov2015constrained,farsoni2018realtime}.
However, these methods still require an explicit parameterization, and their estimates depend on whether the recent trajectory excites the relevant parameters.
TAM has the same basic excitation limitation: if the observed history contains no motion or torque variation that reveals the hidden mismatch, the latent state cannot infer the needed correction.
The difference is that TAM uses a longer robot-side context window, $12\,\mathrm{s}$ in our reported weights, so short stale or low-excitation intervals can still be conditioned on informative motion that occurred earlier in the buffer.
TAM trains this long-horizon Transformer history encoder only through the downstream torque-correction objective, allowing the module to use longer execution context without supervising explicit physical parameters.

\subsection{Action Transformation and Residual Learning}
Another line of work modifies the action sent by a fixed policy rather than changing the policy architecture itself.
Grounded Action Transformation (GAT) grounds a simulator by learning an action transformation inside simulation: when a policy proposes an action, the transformation changes the simulated action so the simulated transition better matches target-domain transitions observed from the physical system~\citep{hanna2017grounded,hanna2021grounded}.
Follow-up variants improve this grounding step by training it with reinforcement learning or by modeling stochastic target dynamics~\citep{karnan2020reinforced,desai2020stochastic}.
Residual reinforcement learning instead keeps an existing controller and learns an additional residual action to improve task reward~\citep{johannink2019residual}.
Recent sim-to-real systems such as ASAP and TRANSIC also introduce correction models that are trained or tuned for a target behavior~\citep{he2025asap,jiang2024transic}.
These methods are close to TAM because they also modify actions or commands to reduce a dynamics mismatch, but the correction is usually tied to a particular policy action space, task objective, or data-collection loop.
As a result, changing the policy may require collecting new grounding data, redefining the residual action, or retraining the correction with a new task reward.
TAM is less sensitive to these choices because it is inserted after the low-level controller: once any policy action is converted to a nominal torque for the ideal robot, TAM uses the same robot-side torque interface and history to output a corrected torque.
Its training data are simulated ideal/perturbed rollouts with teacher-torque labels, rather than real-world grounding data or task reward, so changing the downstream task does not require a new real-world data-collection loop.

\FloatBarrier

\section{TAM Training, Deployment, and Implementation Details}
\label{app:training_randomization}
\label{app:implementation_details}

\subsection{Simulation Data Generation and Randomization}

Table~\ref{tab:training_randomization} lists the randomization ranges used during data generation for the Panda robot.
The first rows specify the hidden physical perturbations applied to the perturbed plant, including rigid-body, payload, actuator, and friction terms.
The final rows list history-side augmentations used during training so that the encoder sees noisy and delayed histories similar to deployment logs.

\paragraph{Rollout generation.}
Training is performed in MuJoCo/MJX simulation~\citep{todorov2012mujoco} using task-agnostic free-space joint-motion rollouts.
Each rollout samples perturbation parameters $\xi \sim p_\Xi$, where $p_\Xi$ is the training distribution over the physics-parameter space $\Xi$, to instantiate a perturbed robot model with varied link mass, inertia, center-of-mass offsets, armature, damping, nonlinear sigmoidal joint friction~\citep{gaz2019dynamic}, end-effector payload, and actuator effects such as dead zone, asymmetric torque bias, and torque scale.
The perturbed robot then executes randomized joint-space waypoint trajectories under a joint impedance low-level controller.
To mimic the effect of object interaction, we inject smooth external generalized torques $w_t$ directly at the joints during these free-space rollouts.
Here $w_t$ approximates the joint-space effect of a contact wrench (i.e.\ $J^\top f$) and is sampled over short, finite time windows; it is not an instantaneous impulse and is not exposed to TAM as an input.
For each rollout, the data generator records $\{q_t,\dot q_t,\tau^a_t,\tau^0_t,w_t,t,\xi\}_{t=1}^{T}$.

\begin{table}[t]
    \centering
    \scriptsize
    \caption{Randomization ranges used during data generation for the Panda robot.}
    \label{tab:training_randomization}
    \setlength{\tabcolsep}{3pt}
    \resizebox{\linewidth}{!}{%
    \begin{tabular}{llll}
        \toprule
        Group & Quantity & Distribution / range & Units / notes \\
        \midrule
        Rigid-body dynamics & Body mass scale & $\mathcal{U}[0.9,1.1]$ & Multiplicative \\
        Rigid-body dynamics & Body COM offset & $\mathcal{U}[-0.01,0.01]$ per axis & m \\
        Rigid-body dynamics & Joint armature & $\mathcal{U}([0.01,\ldots,0.01],[0.5,0.5,0.5,0.5,0.3,0.3,0.3])$ & $\mathrm{kg\,m^2}$ \\
        Payload & EE payload mass & $\mathcal{U}[0,1.5]$ & kg \\
        Payload & EE payload local COM & $\mathcal{U}[-0.075,0.075]$ per axis & m \\
        Actuator & Torque scale & $\mathcal{U}[0.99,1.01]$ per direction & Multiplicative \\
        Actuator & Torque bias & $\tau_b^+\sim\mathcal{U}[-1,1]$, $\tau_b^-=\tau_b^+ + \mathcal{U}[-0.2,0.2]$ & $\mathrm{N\,m}$ \\
        Actuator & Dead zone & $\mathcal{U}[0,1]$ per direction & $\mathrm{N\,m}$ \\
        Actuator & Viscous damping & $\mathcal{U}[0,2]$ per direction & $\mathrm{N\,m\,s/rad}$ \\
        Nonlinear sigmoidal friction & Amplitude & $\mathcal{U}[0.005,3.0]$ per direction & $\mathrm{N\,m}$ \\
        Nonlinear sigmoidal friction & Velocity width & $\mathcal{U}[0.02,0.2]$ per direction & rad/s; converted to sigmoid slope \\
        Nonlinear sigmoidal friction & Velocity shift & $s^+\sim\mathcal{U}[-0.02,0.02]$, $s^-=-s^+ + \mathcal{U}[-0.01,0.01]$ & rad/s \\
        History augmentation & Position noise std. & $\mathcal{U}[10^{-4},10^{-3}]$ & rad \\
        History augmentation & Velocity noise std. & $\mathcal{U}[10^{-3},10^{-2}]$ & rad/s \\
        History augmentation & Velocity-history delay & $\mathcal{U}[0,2]$ & ms; integer $1\,\mathrm{kHz}$ steps \\
        History augmentation & Applied-torque-history delay & $\mathcal{U}[0,4]$ & ms; integer $1\,\mathrm{kHz}$ steps \\
        \bottomrule
    \end{tabular}}
\end{table}

\paragraph{Actuator perturbation model.}
For joint $j$, the nominal torque command is $\tau^0_{t,j}$ and the joint velocity is $\dot q_{t,j}$.
The actuator model includes five perturbations: torque scale, dead zone, torque bias, joint damping, and nonlinear sigmoidal friction.
The torque scale is a constant multiplicative gain within each command direction:
\begin{equation}
    \operatorname{scale}_j(\tau)=
    \begin{cases}
        \alpha_j^+\tau, & \tau\ge0,\\
        \alpha_j^-\tau, & \tau<0.
    \end{cases}
    \label{eq:torque_scale_model}
\end{equation}
The torque bias and damping terms use the velocity direction,
\begin{equation}
    \operatorname{bias}_j(\dot q)=
    \begin{cases}
        b_j^+, & \dot q\ge0,\\
        b_j^-, & \dot q<0,
    \end{cases}
    \qquad
    \operatorname{damp}_j(\dot q)=
    \begin{cases}
        d_j^+\dot q, & \dot q\ge0,\\
        d_j^-\dot q, & \dot q<0.
    \end{cases}
    \label{eq:bias_damping_model}
\end{equation}
The nonlinear sigmoidal friction term also branches by velocity direction:
\begin{equation}
    \begin{aligned}
    \operatorname{fric}_j(\dot q)&=
    \begin{cases}
        A_j^+\!\left[
            \sigma\!\left(k_j^+(\dot q+s_j^+)\right)
            -\sigma\!\left(k_j^+s_j^+\right)
        \right], & \dot q\ge0,\\
        A_j^-\!\left[
            \sigma\!\left(k_j^-(\dot q+s_j^-)\right)
            -\sigma\!\left(k_j^-s_j^-\right)
        \right], & \dot q<0,
    \end{cases}\\
    \sigma(x)=\frac{1}{1+\exp(-x)}.
    \end{aligned}
    \label{eq:actuator_terms}
\end{equation}
Here the subtraction inside each branch makes $\operatorname{fric}_j(0)=0$.
The dead-zone width also branches by input direction, $\delta_j(x)=\delta_j^+$ for $x\ge0$ and $\delta_j^-$ for $x<0$, giving the analytically invertible dead zone
\begin{equation}
    \operatorname{dz}_j(x)=
    \begin{cases}
        \epsilon x, & |x|\le \delta_j(x),\\
        \operatorname{sgn}(x)\left(|x|-\delta_j(x)+\epsilon\delta_j(x)\right), & |x|>\delta_j(x),
    \end{cases}
    \qquad \epsilon=10^{-2}.
    \label{eq:dead_zone_model}
\end{equation}
Evaluating the maps at $\tau^0_{t,j}$ and $\dot q_{t,j}$, the effective torque applied by the perturbed actuator is
\begin{equation}
    \tau^a_{t,j}
    =
    \operatorname{dz}_j\!\left(\operatorname{scale}_j(\tau^0_{t,j})\right)
    +\operatorname{bias}_j(\dot q_{t,j})
    -\operatorname{damp}_j(\dot q_{t,j})
    -\operatorname{fric}_j(\dot q_{t,j}).
    \label{eq:forward_actuator_model}
\end{equation}
At the current state, the velocity-dependent terms are known, so the inverse is analytic.
Given a desired effective torque $\tau^\star_{t,j}$, first remove the additive actuator effects,
\begin{equation}
    y_{t,j}
    =
    \tau^\star_{t,j}
    +\operatorname{damp}_j(\dot q_{t,j})
    +\operatorname{fric}_j(\dot q_{t,j})
    -\operatorname{bias}_j(\dot q_{t,j}).
    \label{eq:actuator_inverse_additive}
\end{equation}
Then compute $z_{t,j}=\operatorname{dz}_j^{-1}(y_{t,j})$ using
\begin{equation}
    \operatorname{dz}_j^{-1}(y_{t,j})
    =
    \begin{cases}
        y_{t,j}/\epsilon, & |y_{t,j}|\le \epsilon\delta_j(y_{t,j}),\\
        \operatorname{sgn}(y_{t,j})\left(|y_{t,j}|-\epsilon\delta_j(y_{t,j})+\delta_j(y_{t,j})\right), & |y_{t,j}|>\epsilon\delta_j(y_{t,j}),
    \end{cases}
    \label{eq:dead_zone_inverse}
\end{equation}
and invert the constant torque scale with
\begin{equation}
    \tau^0_{t,j}
    =
    \begin{cases}
        z_{t,j}/\alpha_j^+, & z_{t,j}\ge0,\\
        z_{t,j}/\alpha_j^-, & z_{t,j}<0.
    \end{cases}
    \label{eq:torque_scale_inverse}
\end{equation}
This per-joint inverse is the state-dependent actuator inverse used inside $B_\xi^\dagger$ in Appendix~\ref{app:inverse_command_map}.

\paragraph{Nominal torque query sampling.}
The waypoint rollout provides the applied torque $\tau^a_t$ at each recorded state.
The nominal torque queries used for teacher-torque supervision are not produced by a downstream policy; they are sampled around the applied rollout torque by adding predefined Gaussian noise,
\begin{equation}
    \tau^0_t
    =
    \tau^a_t + \epsilon_t,
    \qquad
    \epsilon_{t,j}\sim \mathcal{N}(0,\sigma_{\tau,j}^{2}),
\end{equation}
followed by clipping to the training torque range.

\subsection{Deployment Details}
\label{app:deployment_details}

TAM is deployed as a hierarchical system split across two time scales.
Figure~\ref{fig:mam_framework} shows this deployment split between the policy, low-level controller, controller-side torque adaptor, workstation-side history encoder, and target robot plant.
The history encoder uses the GPU: it runs asynchronously on a workstation, consumes the streamed $1\,\mathrm{kHz}$ history buffer with $200$-sample patch stride, and emits an updated latent state $z$ at $5\,\mathrm{Hz}$ for the Franka Panda.
It can run much faster on the workstation GPU, but the long-horizon dynamics context does not need control-rate recomputation, so we use the lower encoder update rate in deployment.
The torque adaptor is the fast component, attached to the low-level controller side and running on the controller CPU at the $1\,\mathrm{kHz}$ hardware rate; at every control tick it receives the most recent latent state, the current robot state, and the nominal torque, applies Equation~\ref{eq:adaptor}, and reuses the latest latent state if a new encoder output has not arrived.
No external-disturbance torque $w_t$ is read from the robot or supplied to TAM at runtime; contact and external loading enter only implicitly through their effect on the proprioceptive and applied-torque history consumed by the encoder.
The two-rate deployment keeps the GPU workload outside the $1\,\mathrm{kHz}$ hardware loop while still using long-horizon temporal context, with the policy and low-level controller unchanged and no policy-specific fine-tuning or dedicated activation trajectory required.

\paragraph{External torque and payload identifiability.}
Robot-model mismatch has two broad sources: actuator-side effects, such as friction and dead zones, and rigid-body effects, such as link or payload mismatch.
Franka's filtered external-torque estimate is formed from measured transmitted joint torque and a rigid-body model; it can reveal link or payload mismatch but can be insensitive to discrepancies between commanded and transmitted torque.
For example, in free space with matched link dynamics, estimated external torque may remain near zero even when friction or a dead zone requires a nonzero command correction.
TAM instead predicts the net correction for actuator-side and rigid-body mismatch from the history $(q,\dot q,\tau^a)$.
Although external torque is not used in our reported experiments, it could serve as an auxiliary cue for distinguishing transient contact from persistent mismatch; evaluating this input is left for future work.

With sufficiently exciting motion and a known actuator model, payload inertial parameters can be identified from joint position, velocity, and torque histories~\citep{swevers1997optimal,farsoni2018realtime}.
When payload and unmodeled actuator effects are both unknown, however, their contributions can be coupled in the same command-to-motion relationship, making unique separation difficult.
TAM does not require this separation: it learns their combined torque correction directly.
Every reported unknown-payload experiment uses the same raw robot-side $(q,\dot q,\tau^a)$ buffer and supplies neither a payload label nor an external-torque measurement; the physics-residual feature $e^\tau$ is computed internally from that buffer.

At an episode or low-level-controller reset, the streaming history cache is cleared and the torque adaptor remains disabled until the history encoder produces an initial latent state; before that point the low-level controller executes the nominal torque directly.
The encoder does not require the full $12\,\mathrm{s}$ horizon to produce a latent state: it decodes patches as soon as enough samples for the first temporal patch and derivative context are available (typically $\approx 0.5\,\mathrm{s}$), with missing or idle rows zeroed and masked.
This is not a separate activation trajectory or excluded warm-up: in reset-condition evaluations the task begins immediately after the reset and the initial pre-enable period remains part of the trial, as stated in Section~\ref{sec:result}.
The carried-state case is used only for the low-level-controller switch tracking experiment in Section~\ref{sec:tracking_experiments}, where the history cache is intentionally preserved across the switch.

\paragraph{Deployment runtime footprint.}
For the Panda weights, the history encoder has about $4.4$M parameters and the torque adaptor has about $63$K parameters.
On an RTX 3080, the measured GPU footprint after encoder/torque adaptor compilation and speed evaluation is about $1.29\,\mathrm{GB}$.
A single-token autoregressive history-encoder update takes $0.999\pm0.045\,\mathrm{ms}$ over $100$ runs (p50 $1.014\,\mathrm{ms}$, p95 $1.042\,\mathrm{ms}$).
This measured update time leaves ample margin for the deployed $5\,\mathrm{Hz}$ encoder rate.
The torque adaptor head takes $0.155\pm0.002\,\mathrm{ms}$ over $200$ runs.

\subsection{Implementation Details}

Table~\ref{tab:implementation_hyperparams} summarizes the runtime and architecture hyperparameters used by the reported TAM weights.
It includes the history horizon, patching parameters, encoder/torque adaptor update rates, cold-start timing, derivative-filtering settings, network dimensions, and torque-supervision settings.
The paragraphs after the table provide extra implementation detail for patch construction, derivative filtering, and training augmentations.

\begin{table}[t]
    \centering
    \caption{Primary TAM implementation hyperparameters. These values are moved out of the main method text to keep the method focused on the algorithmic structure. The cold-start entry describes online deployment timing after a reset, not a separate evaluation warm-up.}
    \label{tab:implementation_hyperparams}
    \resizebox{\linewidth}{!}{%
    \begin{tabular}{lll}
        \toprule
        Component & Hyperparameter & Value \\
        \midrule
        History stream & Nominal history sampling rate & $1\,\mathrm{kHz}$ \\
        History stream & Rolling history horizon & $12\,\mathrm{s}$ \\
        History patching & Patch size / stride & $400$ / $200$ samples \\
        Deployment split & History encoder update rate & $5\,\mathrm{Hz}$ \\
        Deployment split & Torque adaptor control rate & $1\,\mathrm{kHz}$ \\
        Deployment split & Encoder / torque adaptor compute & Workstation GPU / controller CPU \\
        Cold start & First decodable latent state & $\approx 0.5\,\mathrm{s}$ after reset \\
        Derivative filtering & Masked-fit window & $\pm 50$ samples at the nominal history rate \\
        Derivative filtering & Position / velocity equation weights & $2.0$ / $1.0$ \\
        Derivative filtering & Distance taper & $1/(1+|s|)$ \\
        Derivative filtering & Ridge coefficient & $10^{-6}$ \\
        History masking & Idle raw-torque threshold & $10^{-5}$ \\
        History encoder & Causal Transformer layers / heads & $5$ / $8$ \\
        History encoder & Internal token width & $256$ \\
        History encoder & Latent dimension $d_z$ & $64$ \\
        Torque supervision & Nominal torque samples per state $S$ & $64$ \\
        Torque loss & Huber threshold & $10^{-2}$ \\
        \bottomrule
    \end{tabular}}
\end{table}

\paragraph{Patch token construction.}
Let $P$ be the patch length, $R$ the patch stride, and $\ell=1,\ldots,K$ the local index of the rolling buffer in chronological order.
For patch index $m$ and joint $j$, the encoder takes the four-feature temporal patch
\begin{equation}
    \mathcal{P}_{m,j}
    =
    \left[
    q_{\ell,j},\dot q_{\ell,j},\tau^a_{\ell,j},e^\tau_{\ell,j}
    \right]_{\ell=mR+1}^{mR+P}.
    \label{eq:history_patch}
\end{equation}
A patch multilayer perceptron embeds this flattened per-joint sequence into a token $u_{m,j}$, producing tokens indexed by temporal patch $m$ and joint $j$.
The patch tokens are ordered into a causal sequence with learned joint-id embeddings, processed by the causal Transformer stack with rotary positional encoding~\citep{su2021roformer}, and projected to a per-joint latent state $z_{t,j}=E_\phi(h_t)_j\in\mathbb{R}^{d_z}$.

\paragraph{Derivative filtering for the physics residual.}
The acceleration estimate $\ddot{\hat q}_t$ used for $e^\tau_t$ is filtered rather than a raw one-step finite difference.
Around each time $t$, we fit a smooth local curve to nearby position and velocity samples.
Samples closer to $t$ receive more weight, and a small stabilizing penalty keeps the fit well behaved when the local window is noisy or partially masked.
The curvature of this fitted curve gives $\ddot{\hat q}_t$.
Idle or invalid history rows are masked before fitting.
If too few valid samples are available, the implementation falls back to local interpolation or a filtered velocity finite difference, and finally to zero acceleration.
The same masked-fit pipeline is used for real-hardware history reconstruction, so the physics residual is filtered before it reaches the encoder.

\paragraph{Closed-loop rollout loss and training-time augmentations.}
To reduce covariate shift in the short local window, training also includes a short closed-loop rollout loss: starting from the logged window, the perturbed model is stepped under TAM's sampled corrected torque, the resulting $(q,\dot q,\tau^a)$ window is updated, and the same teacher-torque loss is reapplied for the configured rollout length.
The long-horizon latent state $z$ for that segment is still produced from the logged history token.
During training, we additionally apply the history augmentations listed in Table~\ref{tab:training_randomization}: observation noise and randomized input delays for velocity and applied-torque histories.

\FloatBarrier

\section{Torque Retargeting and Explicit-Parameter Baselines}
\label{app:online_sysid_baseline}

\subsection{Inverse Torque-Command Map}
\label{app:inverse_command_map}

For a physics-parameter setting $\xi$, the inverse torque-command map $\Phi_\xi$ maps a desired joint acceleration and external generalized torque to the actuator command required by the corresponding perturbed model:
\begin{equation}
    \Phi_\xi(q_t,\dot q_t,\ddot q_t,w_t)
    =
    B_\xi^{\dagger}\!\left(
        \operatorname{ID}_\xi(q_t,\dot q_t,\ddot q_t)-w_t,
        q_t,\dot q_t\right).
    \label{eq:inverse_command_map}
\end{equation}
Here $\operatorname{ID}_\xi$ is rigid-body inverse dynamics and $B_\xi^\dagger$ is the state-dependent inverse actuator map.
For each sampled physics-parameter setting, $B_\xi^\dagger$ analytically inverts the modeled torque scale, torque bias, dead zone, joint damping, and nonlinear sigmoidal friction terms at the current $(q_t,\dot q_t)$.
The subtraction of $w_t$ means the actuator command supplies only the torque required beyond the known external generalized torque, so by construction the retargeting map preserves the external load $w_t$ in the target motion and does not fold it into the actuator command.
This preservation is a structural property of the label generator, evaluated inside the simulator where $w_t$ is known; we do not separately certify it on the trained network, and at deployment $w_t$ is never measured (see Section~\ref{sec:training_objective}).
In the label generator, the dead-zone branch is given a small nonzero slope, the friction inverse is numerically clipped inside the sigmoid domain, and MuJoCo joint and actuator limits are removed while evaluating $\Phi_\xi$ so that $\tau^\star_{\xi,t}$ represents the corrected torque required before hardware torque limits are applied.
The ideal inverse map $\Phi_0$ is defined analogously with the ideal dynamics and actuator model.

\subsection{Online SysID and PhysPred Baselines}

The online SysID and PhysPred baselines use the same torque-retargeting interface as TAM, but replace TAM's free latent state and torque adaptor with an explicit physical-parameter estimate.
They differ only in how this estimate is obtained.
Online SysID fits $\hat\xi_t$ from recent measured transitions with a particle filter over explicit physical parameters.
PhysPred denotes our controlled learned-parameter baseline, inspired by in-context dynamics-parameter prediction methods such as Dynamics as Prompts~\citep{zhang2025dynamicsprompts}; it uses our explicit parameterization and torque-retargeting pipeline.
PhysPred instead uses TAM's Transformer history encoder to predict $\hat\xi_t$ from history.
Once $\hat\xi_t$ is available, both baselines use the same ideal-acceleration retargeting operation in Equation~\ref{eq:sysid_retarget}.

\paragraph{Online SysID parameter estimation.}
At deployment, it maintains a rolling buffer of measured transitions
\begin{equation}
    \mathcal{D}_t =
    \left\{
    (q_i,\dot q_i,\tau^a_i,q_{i+1},\dot q_{i+1},m_i)
    \right\}_{i=t-N}^{t-1},
\end{equation}
where $\tau^a_i$ is the torque applied to the robot and $m_i\in\{0,1\}$ marks valid consecutive samples.
The reported Online SysID baseline uses a particle-filter implementation with $K=512$ particles, a $50$-sample rolling buffer (49 one-step transitions), and an update stride of $50$ new samples.
The MJWarp scorer evaluates the window in chunks of $10$ transitions, giving an effective parallel batch of $5120$ simulated worlds per chunk.
Before fitting, the real-time wrapper reconstructs a dense sample sequence, removes duplicate timestamps, masks discontinuities, and optionally smooths the local position and velocity traces.

Rather than applying a Kalman filter (KF) or Extended Kalman Filter (EKF) to this nonlinear parameter-estimation problem, the baseline maintains a particle filter over a bounded explicit parameter space $\Xi$ derived from the same randomized-simulation profile used to train TAM.
At each update, $K$ particles $\{\xi^{(k)}\}_{k=1}^{K}$ are scored by the one-step state-prediction error on the current window,
\begin{equation}
    \ell_i\!\left(\xi^{(k)}\right) =
    \left\|
    \begin{bmatrix}
    \sqrt{w_q}\,(\hat q_{i+1}(\xi^{(k)})-q_{i+1}) \\
    \sqrt{w_{\dot q}}\,(\hat{\dot q}_{i+1}(\xi^{(k)})-\dot q_{i+1})
    \end{bmatrix}
    \right\|_2^2 ,
    \label{eq:online_sysid_fit}
\end{equation}
where $(\hat q_{i+1}(\xi),\hat{\dot q}_{i+1}(\xi))$ is the one-step MJX prediction from $(q_i,\dot q_i)$ under the applied torque $\tau^a_i$ and the actuator/dynamics parameters $\xi$.
Particle weights are updated as $w^{(k)} \propto \exp\!\big(-\sum_i m_i\,\ell_i(\xi^{(k)})\big)$, normalized, and used to draw the resampled population for the next window with small parameter-space jitter; the point estimate $\hat\xi_t$ is the weighted mean of the resampled particles, projected back into the bounded parameter ranges.
The optimized parameterization includes the same actuator and rigid-body terms as TAM's training distribution: torque scale, torque bias, dead zone, damping, friction, armature, and a payload-mass term when enabled.
We use a particle filter rather than a KF/EKF estimator because the actuator branches (dead zone, torque bias, sigmoidal friction) make the one-step prediction objective non-smooth and multi-modal in $\xi$, which is not well represented by a single local Gaussian belief.

For all Online SysID entries, we use the same particle-filter hyperparameters unless a table explicitly states otherwise.
The particle count is $K=512$, and the process noise, post-resampling jitter, measurement weights, and resampling threshold are not tuned per downstream task: $w_q=0.1$, $w_{\dot q}=1.0$, measurement temperature $1.0$, process-noise scale $0.01$, and post-resampling jitter scale $0.02$.
The filter resamples when the effective sample size $\mathrm{ESS}=1/\sum_k (w^{(k)})^2$ drops below half the particle count, i.e., $\mathrm{ESS}<0.5K$ with $K=512$ particles.
Each update scores approximately $512\times49$ one-step predictions, batched on GPU through MJWarp; the adaptation process runs outside the hard real-time torque loop and the controller uses the most recent parameter estimate.
The per-control-step retargeting cost after an update is a single ideal-acceleration query and one inverse torque-command map evaluation.

The estimator uses the same ideal robot model, torque limits, proprioceptive state, and applied-torque history available to TAM.
It does not observe ground-truth physical parameters, task rewards, task labels, object state, ball state, future references, or real-world calibration trajectories.
It does receive one advantage that is intrinsic to explicit-parameter baselines: the hand-specified parameter list, bounds, and payload body index.
Where quantities overlap, these bounds are copied from the bounds used to train TAM.

\paragraph{Shared torque retargeting.}
After obtaining $\hat\xi_t$ from either Online SysID or PhysPred, the baseline computes a corrected torque for the current nominal torque $\tau^0_t$ through ideal-acceleration retargeting.
First, it queries the ideal model for the acceleration produced by the nominal command,
\begin{equation}
    \ddot q^0_t =
    F_0(q_t,\dot q_t,\tau^0_t,0).
\end{equation}
Then it evaluates the inverse torque-command map associated with the estimated parameters,
\begin{equation}
    \tau^{\mathrm{sysid}}_t =
    \Phi_{\hat\xi_t}(q_t,\dot q_t,\ddot q^0_t,0),
    \qquad
    \Delta\tau^{\mathrm{sysid}}_t =
    \tau^{\mathrm{sysid}}_t - \tau^0_t .
    \label{eq:sysid_retarget}
\end{equation}
The map $\Phi_{\hat\xi_t}$ uses the same modeled actuator effects as the training label generator, including torque scaling, torque bias, dead zone, damping, and nonlinear sigmoidal friction.
The final torque sent to the robot is clipped to the same jointwise torque limits $[\tau_{\min},\tau_{\max}]$.
Thus the online SysID baseline has access to an explicit parameter-estimation route, but its correction is still constrained by the identified parameter family, the finite history window, and the robot's torque authority.

\section{Tracking Experiment Details}
\label{app:evaluation_protocols}
\label{app:tracking_experiment_details}

\begin{table}[t]
    \centering
    \caption{Experimental taxonomy. This table separates policy type, low-level controller, and metric to avoid conflating tracking, behavior-cloned contact policies, and task execution.}
    \label{tab:experiment_taxonomy}
    \resizebox{\linewidth}{!}{%
    \begin{tabular}{llll}
        \toprule
        Experiment & Policy type & Low-level controller & Metric \\
        \midrule
        Tracking & Reference tracking with unknown parameters & Joint impedance / OSC & Joint or end-effector tracking error \\
        Vision pushing & RL policy & Direct torque & Success rate \\
        Behavior-cloned flip & BC policy & Joint impedance & Success rate \\
        Ball-on-plate & MPC policy & Direct torque & Goals per run \\
        \bottomrule
    \end{tabular}}
\end{table}

Table~\ref{tab:experiment_taxonomy} summarizes the broader experimental structure.
The remainder of this section gives the tracking-specific protocols and cross-robot implementation details.

\subsection{Source-to-OSC Evaluation Protocol}
\label{app:source_to_osc_protocol}

We use a source-to-OSC protocol to evaluate whether adaptation inferred during a simple source motion improves a downstream operational-space control (OSC) segment.
Each trial has two phases.
In the source phase, the robot tracks a prescribed joint-space trajectory with joint impedance control while each adaptation method updates its online state from its allowed robot-side signals.
For TAM, this is the nominal torque and proprioceptive/applied-torque history; for explicit-parameter baselines, this is the transition history used for online parameter estimation.
At a fixed switch time, the source controller is replaced by an OSC target controller that commands an end-effector displacement from the robot's current end-effector pose.
In reset conditions, the adaptation state is cleared at the switch; in carried conditions, the state inferred during the source phase is preserved and used immediately in the OSC phase.
Direct OSC uses the same source and target trajectories but has no adaptation state.

For the Panda experiment, all methods use the same set of source trajectories and target OSC motions.
We log source joint-trajectory RMSE against the ideal source trajectory, target end-effector RMSE against the ideal OSC trajectory, and final end-effector error, which separates source tracking quality from downstream transfer after the switch.
The main paper reports the target-position RMSE because it directly measures the downstream OSC segment.
In the real Panda deployment, the source segment lasts $16\,\mathrm{s}$ and the low-level controller switches to the target OSC segment at $t=16\,\mathrm{s}$; the OSC segment lasts $8\,\mathrm{s}$.
Both source and target references are sampled at $0.01\,\mathrm{s}$.
The source joint reference is a raised-sine-windowed sinusoid with base per-joint amplitudes $(30,25,30,20,30,25,34)^\circ$.
For table evaluations, each rollout randomizes the per-joint amplitude by multiplying the base amplitude by a factor sampled uniformly from $0.75$ to $1.25$, samples each joint's cycle count independently from $3$ to $7$, and randomizes per-joint sign and phase.
The joint-impedance source gains are $K_p=(50,50,50,30,30,30,10)$ and $K_d=(10,10,10,8,8,8,3)$.
The target OSC segment uses five Cartesian waypoints sampled from XYZ bounds $[0.05,0.42]\times[-0.18,0.18]\times[-0.40,0.14]\,\mathrm{m}$ and roll-pitch-yaw bounds $[-40,40]\times[-35,35]\times[-60,60]^\circ$.
The OSC gains are $K_p=(200,200,200,60,60,60)$, $K_d=(30,30,30,10,10,10)$, with nullspace stiffness $0.5$.
Figure~\ref{fig:real_panda_payload_xyz_carried} shows representative carried-state trajectories for the real Panda with the added payload.

\begin{figure}[t]
    \centering
    \includegraphics[width=\linewidth]{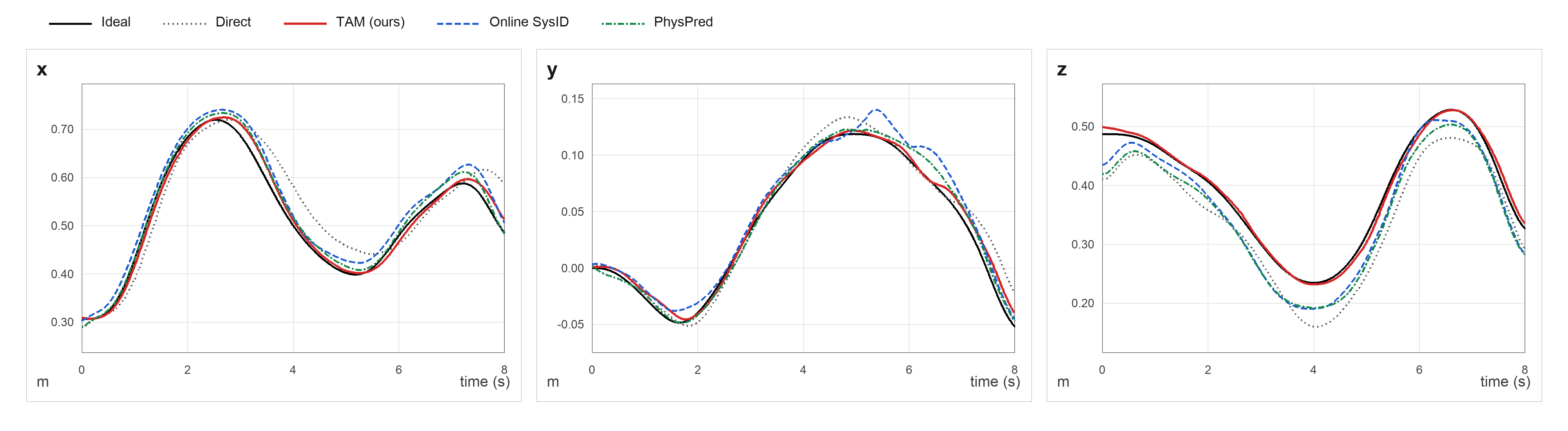}
    \caption{Representative carried-state OSC target trajectories for the real Panda with an added $1\,\mathrm{kg}$ payload, corresponding to the payload setting in Table~\ref{tab:real_ji_to_osc}.
    The plots show end-effector $x$, $y$, and $z$ position coordinates over the $8\,\mathrm{s}$ OSC segment for the ideal-model reference, Direct, TAM, Online SysID, and PhysPred.
    Only carried-state adaptive results are shown.}
    \label{fig:real_panda_payload_xyz_carried}
\end{figure}

\subsection{Cross-Robot Tracking Protocol}
\label{app:cross_robot_tracking_protocol}

For the simulated cross-robot tracking experiment, we instantiate each test robot from its MuJoCo XML asset and robot-specific joint defaults.
Each trial tracks a $16\,\mathrm{s}$ joint-space reference whose sine amplitude and period are sampled independently for each joint.
The multi-robot pretrained weights are trained on Panda, Piper, RBY-1, iiwa14, and Flexiv Rizon4; Google Robot and Unitree Z1 are used only for testing.
The robot-wise simulation fine-tuning baseline starts from these multi-robot weights and then uses target-robot simulation data, representing the practical setting where a robot vendor or developer has the target robot's simulator and can fine-tune TAM without collecting new real-world data.
Simulation metrics are averaged over $100$ randomized reference-trajectory trials and reported as mean $\pm$ standard deviation.

\subsubsection{Cross-Robot TAM Transfer Implementation}
The TAM entries in Table~\ref{tab:cross_robot_tracking} and Table~\ref{tab:cross_robot_etau_ablation} do not train an extra network that maps one robot's joints to another robot's joints.
Instead, each robot is represented directly in its own joint coordinates.
If the test arm has $N$ actuated joints, the controller emits $N$ nominal torques, the history buffer stores $N$ joint histories, the physics-residual feature has $N$ entries, and TAM outputs an $N$-dimensional residual torque.
The history encoder keeps a learned joint-id table with a maximum configured DoF and uses the first $N$ entries at runtime.
Thus 6-DoF robots reuse joint IDs $1$--$6$ by index and 7-DoF robots reuse joint IDs $1$--$7$; no new embedding rows are learned or extended in the zero-shot rows.
Apart from these joint-id embeddings, the patch embedding, Transformer blocks, and torque-adaptor MLP weights are shared across the joint axis.
The torque adaptor applies the same per-joint weights to each joint token and emits an $N$-dimensional residual torque.
Cross-joint information enters through the Transformer history tokens and the torque adaptor's shared global joint summary, rather than through robot-specific output heads.

Torque scaling and limits are handled by the target-robot model.
The nominal torque, applied-torque history, and $e^\tau$ are computed in the target robot's own joint-torque coordinates using its ideal XML model.
The network then uses the normalization associated with the evaluated weights, and the post-TAM command is clipped by the target robot's actuator force ranges before stepping the perturbed plant.
We do not learn a torque-limit normalization or retarget Panda torque limits to the new robot.
This is intentionally a minimal transfer mechanism.
It assumes that the ordered arm joints provide a roughly comparable serial manipulation chain, and it does not perform semantic joint matching, link-level correspondence, or morphology-conditioned embedding lookup.
When joint ordering, actuator semantics, or morphology differ substantially from the training robots, transfer can degrade without additional simulation fine-tuning.
Robots with more joints than the configured joint-id table would require extending the table and retraining or simulation fine-tuning.

\FloatBarrier

\subsubsection{Cross-Robot Physics-Residual Ablation}
The no-$e^\tau$ ablation uses the same multi-robot pretraining setup as TAM, but removes the torque-space discrepancy feature motivated in Section~\ref{sec:method}.
Table~\ref{tab:cross_robot_etau_ablation} extends Table~\ref{tab:cross_robot_tracking} with this ablation and shows that removing $e^\tau$ weakens transfer on every robot, especially the test-only robots, supporting the role of the ideal-model residual in helping the history encoder generalize beyond the robot morphologies seen during multi-robot training.

\begin{table}[!h]
    \centering
    \caption{Extended version of Table~\ref{tab:cross_robot_tracking} with the no-$e^\tau$ ablation added. All values from Table~\ref{tab:cross_robot_tracking} are repeated for comparison. Values are joint-position tracking RMSE in degrees over $100$ randomized simulation trials; lower is better.}
    \label{tab:cross_robot_etau_ablation}
    \scriptsize
    \setlength{\tabcolsep}{3pt}
    \resizebox{\linewidth}{!}{%
    \begin{tabular}{lccccccc}
        \toprule
        Robot & DoF & Split & Direct & Online SysID & TAM w/o $e^\tau$ & TAM & TAM sim-ft. \\
        \midrule
        Panda & $7$ & Train & $4.16 \pm 0.99$ & $1.43 \pm 0.53$ & $1.27 \pm 0.32$ & $\underline{0.59 \pm 0.19}$ & $\mathbf{0.43 \pm 0.17}$ \\
        Piper & $6$ & Train & $4.88 \pm 0.96$ & $0.83 \pm 0.18$ & $0.54 \pm 0.14$ & $\underline{0.52 \pm 0.17}$ & $\mathbf{0.42 \pm 0.15}$ \\
        iiwa14 & $7$ & Train & $3.69 \pm 0.82$ & $0.99 \pm 0.35$ & $0.94 \pm 0.23$ & $\underline{0.56 \pm 0.16}$ & $\mathbf{0.47 \pm 0.16}$ \\
        Flexiv Rizon4 & $7$ & Train & $4.06 \pm 1.20$ & $1.33 \pm 0.51$ & $1.43 \pm 0.74$ & $\underline{0.84 \pm 0.33}$ & $\mathbf{0.68 \pm 0.27}$ \\
        RBY-1 & $7$ & Train & $3.80 \pm 0.81$ & $1.39 \pm 0.83$ & $1.30 \pm 0.23$ & $\underline{0.84 \pm 0.16}$ & $\mathbf{0.66 \pm 0.17}$ \\
        Google Robot & $7$ & Test & $4.69 \pm 1.13$ & $1.65 \pm 0.37$ & $3.58 \pm 1.15$ & $\underline{1.05 \pm 0.40}$ & $\mathbf{0.93 \pm 0.32}$ \\
        Unitree Z1 & $6$ & Test & $6.11 \pm 1.27$ & $1.05 \pm 0.23$ & $2.43 \pm 0.17$ & $\underline{0.93 \pm 0.19}$ & $\mathbf{0.72 \pm 0.24}$ \\
        \midrule
        Mean over robots & -- & -- & $4.48$ & $1.24$ & $1.64$ & $\underline{0.76}$ & $\mathbf{0.62}$ \\
        \bottomrule
    \end{tabular}
    }
\end{table}
\FloatBarrier

\section{Policy Implementation Details}
\label{app:upstream_policy_details}

\subsection{Vision-Based Box Pushing}
\label{app:vision_box_pushing}

\paragraph{Policy training.}
The box-pushing policy uses the push-cube environment in the Sim2realAdaptor codebase, following the MuJoCo Playground-style non-prehensile manipulation setup~\citep{zakka2025mujocoplayground}.
The task is to push a box on the table to a commanded planar goal pose.
We train the policy in two stages.
First, a state-based teacher policy is trained in simulation with PPO.
The teacher observes privileged simulator state, including robot joint position and velocity, end-effector pose, box pose, target pose, and previous action.
Its reward combines terms for approaching the box from the side, reducing box-to-target position error, reducing box orientation error, maintaining useful side contact, and regularizing robot motion and action changes.

\paragraph{Point-cloud student.}
The deployed vision policy is trained by DAgger from the state-based teacher rather than by RL directly from point clouds.
At each DAgger step, the rollout state is observed by the teacher, the teacher action is treated as the supervision target, and the student is optimized to match that joint-torque action.
The student receives two partial depth point clouds rendered from randomized cameras.
Both clouds are transformed into the robot-base coordinate frame and concatenated into a fused point set, where each point contains $(x,y,z)$ and a one-hot source label.
Real camera points use view labels, and the fused point set also includes two synthetic geometry groups: gripper augmentation adds $20$ surface keypoints sampled on the Panda hand and fingers from the current robot state, and goal augmentation adds the target-box mesh vertices transformed to the commanded goal pose.
The fused point set is passed to the policy together with non-visual inputs such as robot proprioception and the commanded goal.
These synthetic gripper and goal vertices give the PointNet encoder explicit geometric anchors for the contact tool and the desired object pose, while the real camera points carry the observed object and scene geometry.
The visual backbone is a PointNet-style encoder~\citep{qi2017pointnet}: a shared per-point MLP embeds the fused point cloud, masked mean pooling produces a global point feature, and a recurrent GRU head combines this feature with the scalar inputs to output the joint-torque action.
The student also has an auxiliary box-vertex prediction head trained from simulator geometry, which encourages the point-cloud representation to preserve object pose information useful for pushing.

\paragraph{Real-world box-pose estimation and point-cloud preprocessing.}
For real-world pushing, the deployment stack does not read simulator object state.
Depth observations are converted to a point cloud in the robot-base frame using the calibrated camera pose.
Before either policy inference or box-pose tracking, the cloud is cropped to the calibrated image region, projected with the deployment intrinsics, filtered to the robot workspace, and stripped of table points.
When a calibration table plane is available, deployment uses that plane directly.
Otherwise, the policy-side preprocessor fits a table plane from the current cloud with a RANSAC-style procedure: it samples candidate points, evaluates random triplet planes, refines the best inlier plane, and removes inliers within a $1\,\mathrm{cm}$ tolerance.
The deployed point-cloud policy keeps at most $6000$ non-table camera points per view after filtering.

The box pose used for real-world state estimation and success evaluation is estimated from the filtered object cloud with a sequential particle filter~\citep{thrun2005probabilistic}.
The filter state is the planar box pose $(x,y,\psi)$ on the table; the vertical position is fixed by the estimated table height plus the known object support offset.
Particles are initialized around the previous valid pose when available, otherwise around the median observed object position, and are propagated with a random-walk model.
Each particle is scored by a point-to-mesh likelihood between the known box mesh and the observed object points, using clipped point-to-surface distances and an inlier count.
In the reported deployment, the filter maintains $256$ candidate box poses.
The distance scale in the likelihood is $0.005\,\mathrm{m}$: particles whose transformed box mesh lies within a few millimeters of the observed object points receive much higher weight.
After each update, the filter resamples if the effective number of particles falls below $128$, so low-weight pose hypotheses are replaced by samples near high-weight ones.
The previous valid pose is reused only when it is recent; cached poses older than $0.25\,\mathrm{s}$ are discarded and the filter reinitializes from the current object cloud.
For success evaluation, a real-world pushing trial is counted successful when this estimated planar box pose is within $0.03\,\mathrm{m}$ of the target position and within $15^\circ$ of the target yaw before the $15\,\mathrm{s}$ timeout.

\paragraph{RMA policy-side baseline.}
The RMA pushing baseline uses the same teacher-student structure but adds policy-side adaptation.
The RMA teacher is trained with a privileged physics encoder that maps the simulator's randomized physical parameters to a latent state used by the state-based policy.
During DAgger, the vision student learns an additional history-to-latent-state module from recent scalar observations and past actions.
This module is supervised to predict the teacher's privileged physics latent state, and the predicted latent state conditions the student policy.
Thus the RMA baseline adapts inside the pushing policy, while TAM is inserted after the policy at the torque interface and does not use the policy's privileged latent state or reward.
In the real-world pushing runs reported in Table~\ref{tab:realworld_stats}, adding TAM improves both the native policy and the RMA policy, but the native-policy TAM result is higher than TAM with RMA.
Rollout diagnostics suggest that the dominant difference is policy-side: the RMA policy is trained with stronger dynamics randomization and produces a more conservative pushing strategy.
The RMA policy shows a smaller accumulated gripper-orientation change than the native policy ($250.23^\circ$ vs.\ $334.11^\circ$), supporting the interpretation that RMA reduces aggressive wrist motion during pushing.

\FloatBarrier

\subsection{Behavior-Cloned Flip Policy}
\label{app:bc_flip_details}

The flip demonstrations are generated in ideal-model MuJoCo simulation with a Franka robot rather than collected from human teleoperation.
A scripted demonstrator produces the flip rollouts directly in simulation: it moves the gripper to a pre-flip pose, executes a contact-rich reorientation motion that flips the block, and then retracts.
We run these scripted rollouts with cameras enabled and merge the successful episodes into a LeRobot-format training dataset.

The recorder stores RGB/depth camera streams, robot state, object poses, applied torques, and controller targets.
For the ACT policy used in the real-world flip experiment, we train on the gripper-camera RGB stream and the compact robot state, with action targets consisting of seven absolute joint targets and a gripper command.

The policy is trained with the LeRobot ACT training pipeline using the standard ACT architecture~\citep{zhao2023act}.
In the inspected deployment weights, the visual encoder is a ResNet-18 backbone, the policy uses a Transformer encoder--decoder with a conditional VAE latent variable, and it predicts a chunk of future joint-target and gripper actions.
At deployment, the predicted joint targets are tracked by the joint-impedance controller.
TAM is not part of ACT training; it is inserted afterward at the torque interface and corrects the nominal torque produced by the joint-impedance controller.
Real-world success is judged from the final object orientation: the trial succeeds when the desired cube face is on the bottom after the flip episode.

\subsection{Ball-on-Plate MPC}
\label{app:ball_plate_mpc_details}

The ball-on-plate policy is implemented as a torque-space wrapper around the MuJoCo MPC Franka ball-balance task~\citep{howell2022predictive}.
At each control step, the wrapper writes the commanded ball goal into the MuJoCo MPC task userdata, injects the measured arm state and estimated ball state into the planning model, runs the MPC agent, and reads the first torque action from the planned trajectory as the nominal torque.
The real deployment uses a $20\,\mathrm{ms}$ MPC agent timestep, a $0.4\,\mathrm{s}$ planning horizon, and a $50\,\mathrm{Hz}$ control loop.
TAM is inserted after this MPC torque output, so the iLQR planner and ball-state objective are unchanged.
Goal positions are sampled uniformly in area over a radial range of $0.02$--$0.05\,\mathrm{m}$ in the plate frame, clipped to the valid plate region.
A goal-reaching event is counted when the estimated planar ball position enters a $5\,\mathrm{mm}$ radius around the current target; a $10\,\mathrm{mm}$ exit threshold is used as hysteresis before another reach event can be counted.
When a goal is reached, a new target is immediately sampled and the episode continues until the $15\,\mathrm{s}$ timeout.

\paragraph{iLQR residual design.}
The MuJoCo MPC task is written as a weighted residual minimization problem.
At the beginning of each MPC solve, the task constructs a short plate-frame ball reference from the measured ball position to the commanded goal.
The reference evolves with
\begin{equation}
    v^{\mathrm{ref}}_k =
    \operatorname{clip}\!\left(k_g(g - p^{\mathrm{ref}}_k), v_{\max}\right),
    \qquad
    p^{\mathrm{ref}}_{k+1} = p^{\mathrm{ref}}_k + \Delta t\, v^{\mathrm{ref}}_k ,
\end{equation}
where $g$ is the target ball position on the plate.
In the deployed configuration, $k_g=1.0$, $v_{\max}=0.04\,\mathrm{m/s}$, and $\Delta t=20\,\mathrm{ms}$.
This produces a smooth local reference over the MPC horizon instead of asking iLQR to jump directly to the goal.

The residual vector has six blocks.
The primary task residual is the two-dimensional plate-frame ball error, $p^{\mathrm{ball}}_k-p^{\mathrm{ref}}_k$.
The remaining residuals regularize the robot and keep the ball in a valid region: plate-center displacement from the home pose, joint-posture deviation from the home configuration, hinge losses on joint-velocity limits, hinge losses on torque effort, and a hinge loss when the ball moves outside the valid plate region.
The corresponding task weights in the deployed XML are $40.0$ for ball tracking, $1.0$ for plate-center position, $0.1$ for joint posture, $5.0$ for joint-velocity limits, $0.2$ for torque effort, and $1000.0$ for the ball-position boundary.
Thus the iLQR planner prioritizes moving the ball along the short-horizon reference while softly discouraging aggressive arm motion and strongly penalizing losing the ball from the usable plate area.

\paragraph{Analytical ball-position estimator.}
The reported real-world ball-on-plate experiments use an analytical estimator rather than a learned image estimator.
At startup, the ball is removed from the plate and the system captures eight RGB calibration frames of the checkerboard.
The calibration routine median-stacks these frames, reads camera intrinsics from the calibrated camera stream or a JSON file, detects the checkerboard grid, and fits a homography from known plate-frame checker points to image pixels.
Using the calibrated camera intrinsics, this homography is decomposed to recover the camera-to-plate transform.
The default geometry assumes a $0.18\,\mathrm{m}$ square plate, a $4\times4$ checker grid, and a $0.045\,\mathrm{m}$ diameter ball.

During each deployment step, the red ball is segmented in HSV space using fixed hue, saturation, and value thresholds.
The largest valid connected component gives the ball centroid in image pixels.
Given the known ball radius and calibrated plate geometry, the estimator casts a camera ray through the centroid and intersects it with the ball-center plane offset from the plate surface by the ball radius.
This produces a two-dimensional ball position in the plate frame.
A deployment state filter estimates velocity from recent position estimates before passing the ball state to the MPC planner.
If segmentation or ray intersection fails for a frame, the runtime falls back to the last valid ball position.

\FloatBarrier

\section{Additional Analyses and Ablations}
\label{app:additional_analyses}

\subsection{Architecture Ablations}
\label{app:architecture_ablations}

We run a simulation design ablation on the push-policy task to test which architectural choices matter.
Table~\ref{tab:architecture_ablations} evaluates adaptation variants in simulation using $100$ push-policy trials.
\begin{table}[!h]
    \centering
    \scriptsize
    \caption{Simulation ablations for TAM architecture choices and explicit-physics baselines.
    Component columns mark whether each variant uses the long-horizon history encoder, torque adaptor, and physics residual feature $e^\tau$.
    The adaptation-target column distinguishes TAM's direct torque-correction objective from explicit-parameter and privileged-latent-state baselines.
    Success is measured over $100$ trials.
    We report success rate only.}
    \label{tab:architecture_ablations}
    \setlength{\tabcolsep}{3.5pt}
    \resizebox{\linewidth}{!}{%
    \begin{tabular}{lccclc}
        \toprule
        \multirow{2}{*}{Variant} & \multicolumn{3}{c}{Component present} & \multirow{2}{*}{Adaptation target} & \multirow{2}{*}{Success rate} \\
        \cmidrule(lr){2-4}
        & Hist. enc. & Torque adaptor & $e^\tau$ & & \\
        \midrule
        Direct transfer & \no & \no & \no & None & $21\%$ ($21/100$) \\
        Short-window torque adaptor only & \no & \yes & \no & Direct torque correction & $42\%$ ($42/100$) \\
        No physics residual feature & \yes & \yes & \no & Direct torque correction & $67\%$ ($67/100$) \\
        History predicts physics parameters & \yes & \no & \yes & Explicit physical parameters & $59\%$ ($59/100$) \\
        Privileged-parameter latent-state supervision & \yes & \yes & \yes & Oracle physical-parameter latent state & $70\%$ ($70/100$) \\
        Full TAM & \yes & \yes & \yes & Direct torque correction & $\mathbf{84\%}$ ($84/100$) \\
        \bottomrule
    \end{tabular}
    }
\end{table}
\FloatBarrier
\textbf{Direct transfer} sends the nominal torque to the robot without adaptation.
\textbf{Full TAM} uses the history encoder, physics residual feature, and torque adaptor.
\textbf{Short-window torque adaptor only} removes the history encoder and gives the torque adaptor only a recent window of state and torque history.
\textbf{No physics residual feature} removes $e^\tau$, the ideal-model inverse-command residual computed from measured motion, while keeping the same raw history stream and Transformer capacity.
\textbf{History predicts physics parameters} predicts explicit physical parameters from history and then applies the shared torque-retargeting operation in Appendix~\ref{app:online_sysid_baseline}.
\textbf{Privileged-parameter latent-state supervision} introduces a comparison-only training target inspired by RMA-style privileged adaptation: a physical-parameter encoder maps the simulator's oracle physical parameters to a latent state $z$, the history encoder is supervised to predict that latent state, and the torque adaptor decodes from it.
Full TAM reaches $84\%$ success, compared with $21\%$ for direct transfer, $42\%$ without the history encoder, $67\%$ without $e^\tau$, $59\%$ for explicit parameter prediction, and $70\%$ for privileged-parameter latent-state supervision.
This privileged-latent-state row is not a component removed from Full TAM; it is a baseline that tests whether supervising the latent state toward oracle physical parameters is better than learning the latent state only through the torque-correction objective.
These results support three separate design claims: long-horizon history matters, the physics residual feature matters, and direct residual-torque decoding outperforms forcing the adaptation state through explicit physical-parameter prediction or privileged-parameter latent-state supervision in this setting.

\FloatBarrier

\subsection{Latent-State Bottleneck Analysis}
\label{app:bottleneck_metrics}

The bottleneck analysis asks whether the latent state organizes histories by correction-relevant physical condition rather than simply memorizing trajectory identity.
For Table~\ref{tab:latent_bottleneck}, we use a crossed held-out bank with $C=64$ sampled physical conditions and $A=64$ fixed push-policy trajectories.
Each trajectory ID is generated by rolling out the push policy once, taking the realized arm $(q,\dot q)$ trace as the reference, resampling it to the torque adaptor timestep, and estimating $\ddot q$ from the resampled trace.
For every pair $(\xi_c,T_a)$, the same reference trajectory is replayed under condition $\xi_c$ using inverse-dynamics feed-forward plus high-PD feedback with gains $k_p=100$ and $k_d=20$.
This produces a controlled bank of histories $h_{c,a}$ in which trajectory identity and physical condition are crossed.
The replay sanity check has $q$-tracking RMSE $1.12\times 10^{-4}$, so the fixed trajectory IDs are followed closely across sampled conditions.
All rows in Table~\ref{tab:latent_bottleneck} use training step $150{,}000$ and the same held-out bank.

\paragraph{Metric computation.}
We report three metrics: train loss, held-out $\tau$ RMSE, and the latent-state invariance ratio $R_z$.
The held-out torque error evaluates whether the latent state supports the downstream correction map.
For each history, the evaluator encodes the first half of the replay and evaluates the torque adaptor on valid timesteps in the second half.
The prediction is the residual torque $\Delta\tau_\theta$ output by the torque adaptor, and the target is the teacher residual
\begin{equation}
    \Delta\tau^\star_{\xi,t}
    =
    \Phi_\xi(q_t,\dot q_t,F_0(q_t,\dot q_t,\tau^0_t,0),0) - \tau^0_t .
\end{equation}
The held-out $\tau$ RMSE is the root mean-squared error between $\Delta\tau_\theta$ and $\Delta\tau^\star_{\xi,t}$ over joints, valid timesteps, trajectories, and physical conditions.

The latent-state ratio $R_z$ measures whether the history encoder groups states by physical condition instead of by trajectory identity.
For each replay history, the history encoder produces a latent state $z_{c,a}=E_\phi(h_{c,a})$.
Latent states are first diagonal-whitened by subtracting the global mean and dividing each latent-state coordinate by its global standard deviation.
For two whitened latent states $\tilde z_i$ and $\tilde z_j$, the distance is
\begin{equation}
    \delta_z(i,j)
    =
    \frac{\|\tilde z_i-\tilde z_j\|_2}{\sqrt{d_z}} .
\end{equation}
The invariance distance compares different trajectories under the same physical condition,
\begin{equation}
    D_z(\mathrm{same}\ \xi,\mathrm{diff}\ T)
    =
    \mathbb{E}_{c,\ a\ne b}
    \left[
    \delta_z\big((c,a),(c,b)\big)
    \right],
\end{equation}
while the separation distance compares different physical conditions under the same trajectory,
\begin{equation}
    D_z(\mathrm{diff}\ \xi,\mathrm{same}\ T)
    =
    \mathbb{E}_{c\ne d,\ a}
    \left[
    \delta_z\big((c,a),(d,a)\big)
    \right].
\end{equation}
The reported ratio is
\begin{equation}
    R_z =
    \frac{
    D_z(\mathrm{same}\ \xi,\mathrm{diff}\ T)
    }{
    D_z(\mathrm{diff}\ \xi,\mathrm{same}\ T)
    } .
\end{equation}
Lower $R_z$ means that histories from the same physical condition but different trajectories are closer than histories from different physical conditions under the same trajectory.

\begin{table}[!htbp]
    \centering
    \footnotesize
    \caption{Bottleneck dimension and motion-level invariance on the push-policy crossed bank ($C=64$ physical conditions, $A=64$ fixed trajectories). $R_z$ is the latent-state invariance/separation ratio $D_z(\mathrm{same}\ \xi,\mathrm{diff}\ T) / D_z(\mathrm{diff}\ \xi,\mathrm{same}\ T)$. Lower is better for all columns. Train loss is averaged from logged evaluations around training step $150{,}000$; held-out $\tau$ RMSE is the offline residual-torque prediction error.}
    \label{tab:latent_bottleneck}
    \begin{tabular}{lccc}
        \toprule
        $d_z$ & Train loss $\downarrow$ & Held-out $\tau$ RMSE $\downarrow$ & $R_z$ $\downarrow$ \\
        \midrule
        $16$ & $0.822$ & $1.038$ & $0.200$ \\
        $64$ & $0.784$ & $0.977$ & $0.213$ \\
        $256$ & $0.777$ & $1.017$ & $0.341$ \\
        \bottomrule
    \end{tabular}
\end{table}

\FloatBarrier

Table~\ref{tab:latent_bottleneck} shows that, among the three sizes we tested, the $64$D bottleneck is the best functional tradeoff: too small (16D) underfits, with worse held-out torque RMSE despite a marginally lower $R_z$; too large (256D) lowers training loss but admits trajectory information into the latent state, raising $R_z$ from $0.213$ to $0.341$ without improving held-out correction.
The 16D-vs-64D held-out RMSE gap ($1.038$ vs $0.977$) is small, so we read this as evidence that capacity matters but that the precise value is not sharp; we report it as ``best in this sweep'' rather than a globally tuned optimum.

\FloatBarrier

\begin{figure}[t]
    \centering
    \begin{minipage}{0.49\linewidth}
        \centering
        \includegraphics[width=\linewidth]{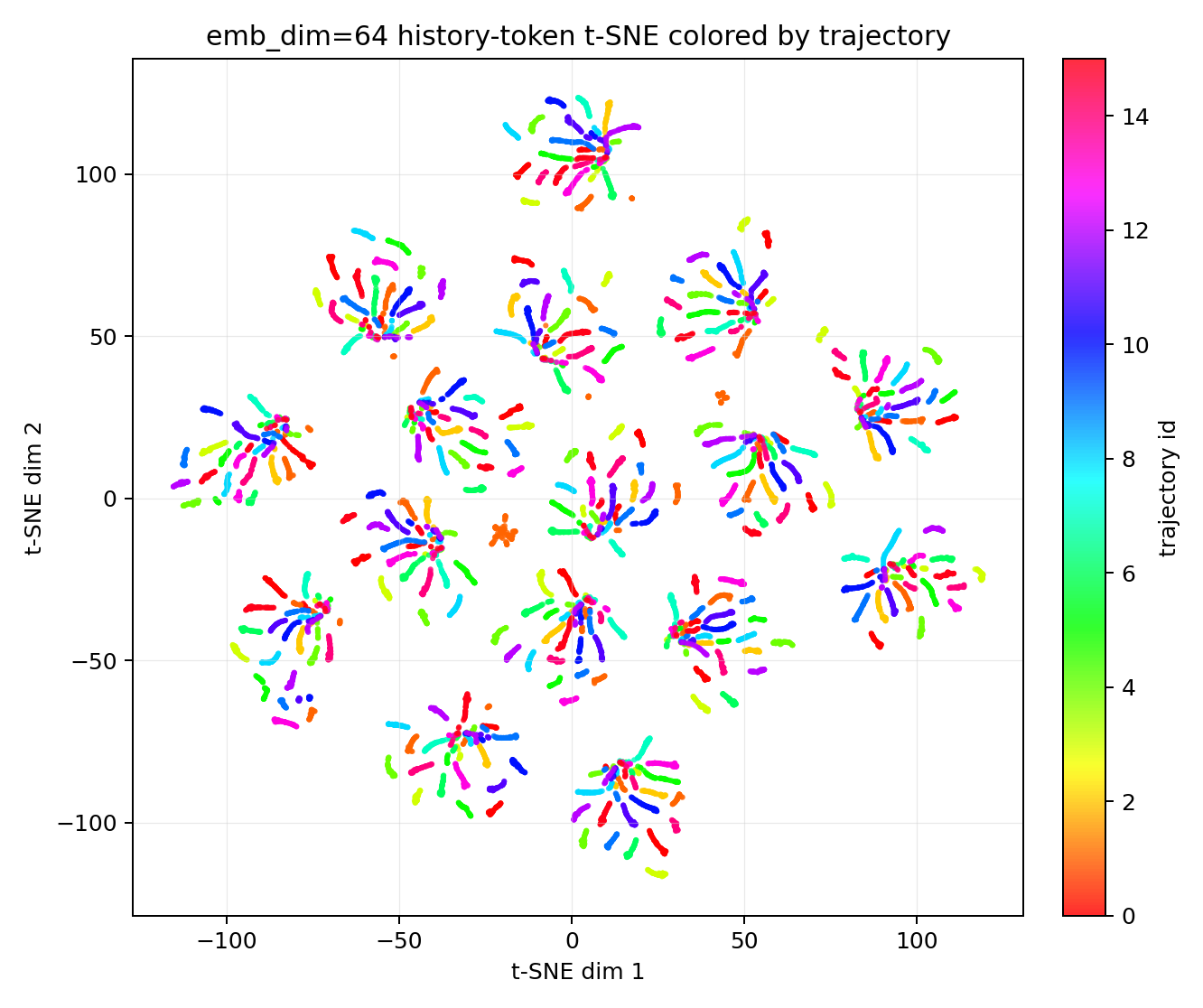}\\[-0.5ex]
        \footnotesize (a) Colored by trajectory ID.
    \end{minipage}
    \hfill
    \begin{minipage}{0.49\linewidth}
        \centering
        \includegraphics[width=\linewidth]{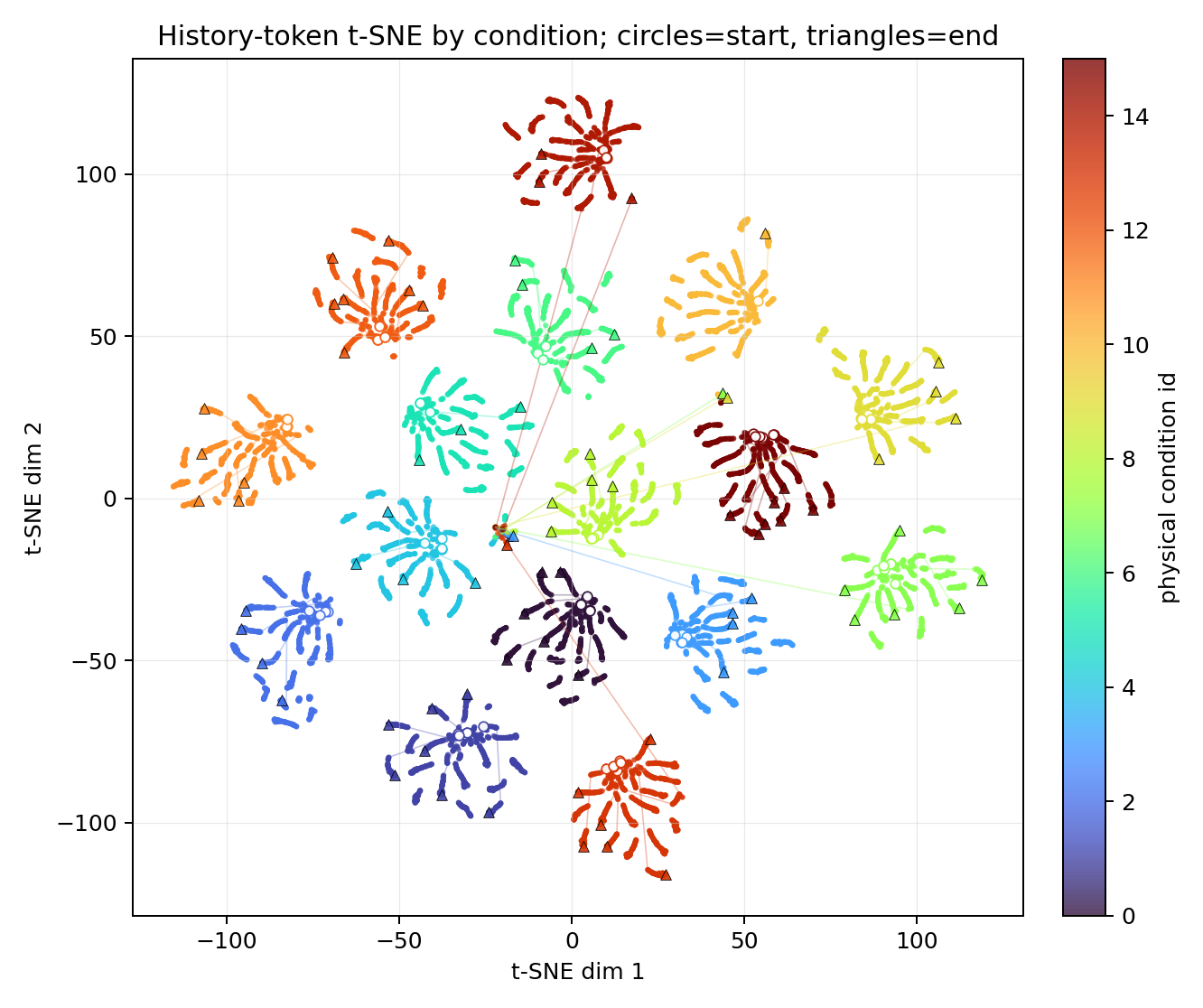}\\[-0.5ex]
        \footnotesize (b) Colored by physical condition ID.
    \end{minipage}
    \caption{Appendix t-SNE visualization of history-token embeddings for the $64$D bottleneck model.
    The same embeddings are colored by trajectory ID in (a) and physical condition ID in (b).}
    \label{fig:appendix_tsne_history_tokens}
\end{figure}

\FloatBarrier

\subsection{No-Mismatch Check: TAM on the Ideal Robot}
\label{app:no_mismatch_check}

A well-behaved adaptation layer should not degrade a policy that already transfers successfully.
To test whether TAM can degrade such a policy, we apply it to the exact ideal robot in simulation: there the model mismatch is zero, the correct residual is zero, and any TAM-induced change in behavior would be a degradation.
Task-level performance is unchanged: the box-pushing policy succeeds in $84\%$ of $100$ trials both with and without TAM, and the MPC ball-on-plate controller completes $15.9$ goals per $15$ s in both cases ($20$ runs each), matching the ideal-simulation reference in Table~\ref{tab:realworld_stats}.
At the motion level, Figure~\ref{fig:app_ideal_plant_tracking} compares TAM-enabled execution against the ideal rollout on a representative randomized $16$ s source trajectory: TAM's rollout follows the ideal rollout with a joint-position RMSE of $0.20\pm0.01^\circ$ across five rollouts, and its residual torque stays below $0.91$ Nm on every joint, far below the actuator torque limits.
We observed no measurable degradation in these tested no-mismatch settings.

\begin{figure}[!t]
  \centering
  \includegraphics[width=0.82\linewidth]{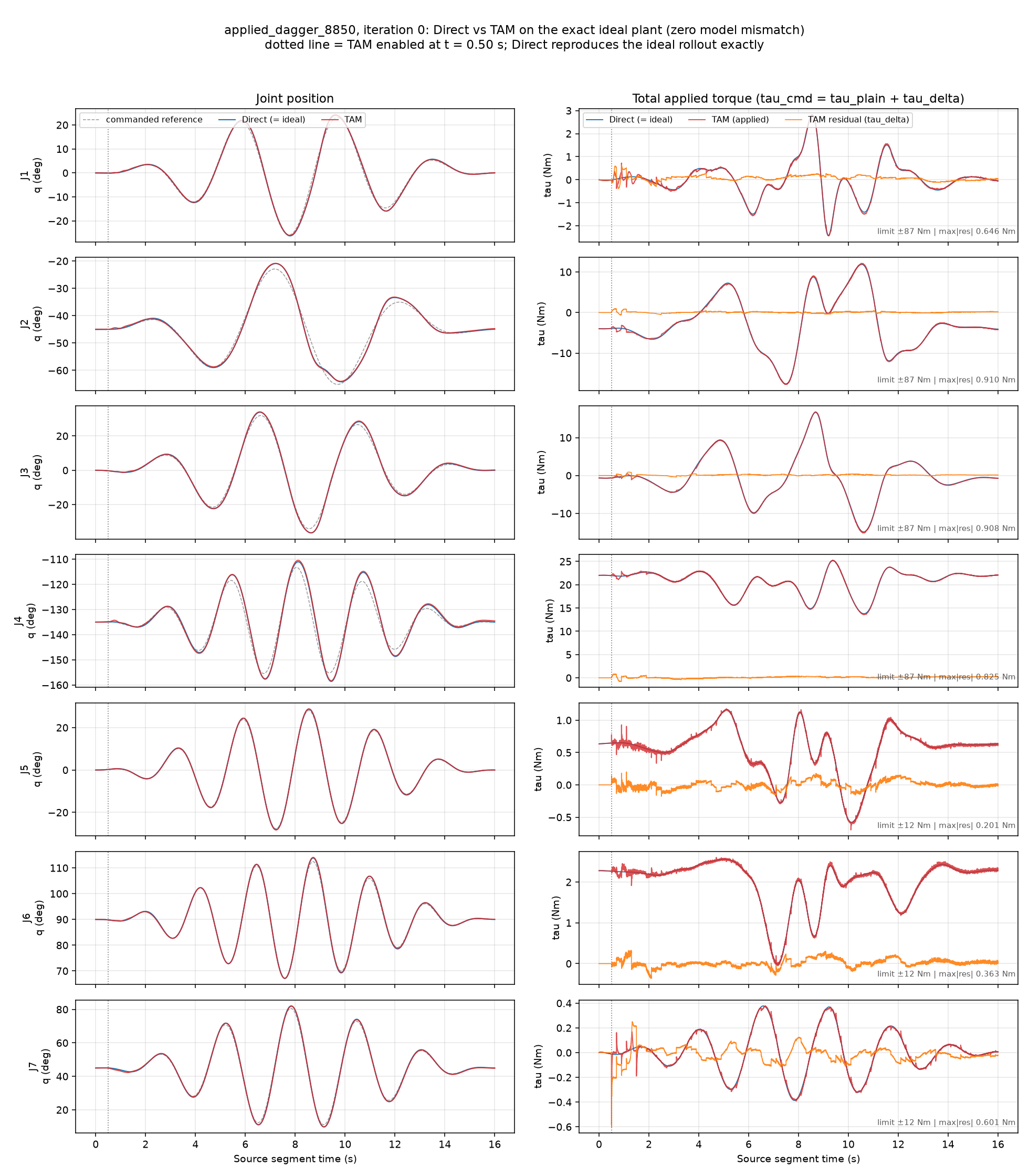}
  \caption{Direct versus TAM execution on the exact ideal plant (zero model
  mismatch) for one representative $16$ s randomized source trajectory.
  Left column: joint positions; the Direct rollout reproduces the ideal
  rollout exactly, and the TAM rollout is visually indistinguishable from it.
  Right column: total applied torque and TAM's residual torque; the per-joint
  peak residual (annotated in each panel) stays below $0.91$ Nm, far below
  the joint torque limits. The dotted line marks TAM's first latent update at
  $t=0.5$ s.}
  \label{fig:app_ideal_plant_tracking}
\end{figure}

\FloatBarrier

\subsection{Out-of-Distribution Friction Robustness}
\label{app:ood_friction}

TAM's robustness claim is empirical within the randomized mismatch family used during training and the available torque authority, not universal robustness.
To probe conditions beyond the training bounds directly, we artificially increase the target robot's joint friction to $2\times$ and $3\times$ the maximum value covered by the randomized training distribution and measure box-pushing success in simulation ($100$ trials per condition).
Table~\ref{tab:ood_friction} reports the results; the no-mismatch row repeats the ideal-robot reference from Appendix~\ref{app:no_mismatch_check}.

\begin{table}[!h]
    \centering
    \small
    \caption{Box-pushing success in simulation under out-of-distribution joint
    friction. Friction is scaled relative to the maximum bound of the
    randomized training distribution; each entry uses $100$ trials.}
    \label{tab:ood_friction}
    \begin{tabular}{lcc}
        \toprule
        Joint friction condition & TAM & Direct \\
        \midrule
        Ideal model (no mismatch) & $84\%$ & $84\%$ \\
        $2\times$ maximum training bound & $\mathbf{82\%}$ & $75\%$ \\
        $3\times$ maximum training bound & $\mathbf{80\%}$ & $59\%$ \\
        \bottomrule
    \end{tabular}
\end{table}
\FloatBarrier

TAM degrades gracefully under friction beyond its training distribution, while direct transfer deteriorates sharply at $3\times$ the training bound.
Broader shifts can be covered by expanding the simulation randomization range and fine-tuning TAM in simulation, which requires no real-robot data.
Regarding out-of-distribution contact patterns, TAM training applies only random perturbations to the robot and never observes the contact patterns of box pushing or flipping; the real-task gains over direct transfer in Table~\ref{tab:realworld_stats} therefore provide empirical evidence of robustness to contact patterns unseen during training.

\FloatBarrier

\subsection{Adaptation-Latency Protocol and Warm Start}
\label{app:adaptation_latency}

Before its first latent state is available ($\approx0.5$ s), TAM passes the nominal torque through unchanged, so tasks that require adaptation from the very first control tick remain a limitation.
Without an abrupt dynamics change, tracking error falls by $90\%$ after this $0.5$ s warm start.
The abrupt-change experiment in Section~\ref{sec:adaptation_latency} uses $20$ random trajectories, each starting $5$ s before the physical parameters switch at $t=0$.
We sample random joint targets and interpolate them with a smooth spline to generate each trajectory.
We track joint-position RMSE and measure recovery relative to the original TAM's steady-state error level ($1.83^\circ$), using the same reference level for all variants.
This recovery time includes closed-loop trajectory catch-up after the switch and should not be interpreted as the history encoder's computation time or parameter-identification latency alone.
The variants are defined in Section~\ref{sec:dagger_fusion}, and Figure~\ref{fig:adaptation_latency} presents their comparison.

\FloatBarrier

\section{Stability and Oscillation Analysis}
\label{app:stability_oscillation}

TAM adds a learned residual torque below the low-level controller, so a natural concern is whether this residual can destabilize the closed loop or excite oscillations that the upstream policy never commanded.
We do not claim a general stability or safety guarantee.
Instead, this section provides two complementary checks on TAM's closed-loop behavior: a conditional boundedness result showing that joint-position and joint-velocity tracking errors remain bounded for all future times under the stated controller and disturbance assumptions (Appendix~\ref{app:boundedness_derivation}), and an empirical frequency-domain analysis of real-Panda executions showing no dominant TAM-induced oscillation peak in the tested trajectories (Appendix~\ref{app:oscillation_spectrum}).

\subsection{Bounded Tracking Errors Under a Bounded Residual}
\label{app:boundedness_derivation}

We apply the standard manipulator Lyapunov argument~\citep{spong2020robot}, treating TAM's induced torque as a bounded perturbation.
The idea is an energy balance: damping removes tracking-error energy quadratically in the error magnitude, whereas a bounded torque can supply energy at most linearly.
Once the error is sufficiently large, damping dominates and the error energy must decrease.
We use this observation to establish a fixed energy ceiling and hence bound both position and velocity tracking errors.

\paragraph{Tracking errors.}
Let $N\in\mathbb{N}_{>0}$ be the number of actuated joints, $t\geq0$ denote time, and $q(t),q_d(t)\in\mathbb{R}^N$ denote the actual and smooth reference joint positions.
Dots denote time derivatives.
Define the joint-position error, joint-velocity error, and filtered tracking error as
\begin{equation}
  e=q-q_d,\qquad
  \dot e=\dot q-\dot q_d,\qquad
  s=\dot e+\Lambda e,
  \label{eq:app_error_definition}
\end{equation}
where $e,\dot e,s\in\mathbb{R}^N$ and the constant filter gain $\Lambda\in\mathbb{R}^{N\times N}$ is symmetric positive definite.
The filtered error combines position and velocity error and gives the stable first-order equation
\begin{equation}
  \dot e=-\Lambda e+s.
  \label{eq:app_error_filter}
\end{equation}
The term $-\Lambda e$ drives the position error toward zero, while $s$ acts as its input.
We first bound $s$, then use this equation to bound $e$ and $\dot e$.
Here $e$ is distinct from the physics-residual feature $e^\tau$ used elsewhere in the paper.

\paragraph{Robot dynamics and controller.}
Write the robot dynamics relative to the rigid-body model used by the base controller as
\begin{equation*}
  M(q)\ddot q+C(q,\dot q)\dot q+g(q)=\tau+d_0.
\end{equation*}
Here $M(q)\in\mathbb{R}^{N\times N}$ is the model's symmetric positive-definite inertia matrix, $C(q,\dot q)\in\mathbb{R}^{N\times N}$ its Coriolis/centrifugal matrix, and $g(q)\in\mathbb{R}^N$ its gravity torque.
The input $\tau\in\mathbb{R}^N$ is a joint torque, while $d_0(t)\in\mathbb{R}^N$ collects the remaining model/actuator errors and disturbances as a joint torque.

Define the reference velocity and its derivative by
\begin{equation*}
  \dot q_r=\dot q_d-\Lambda e,\qquad
  \ddot q_r=\ddot q_d-\Lambda\dot e.
\end{equation*}
Then $s=\dot q-\dot q_r$ and $\dot s=\ddot q-\ddot q_r$.
Consider a base controller that supplies the modeled reference-motion torque and damps the filtered error, with TAM adding its residual:
\begin{equation*}
  \tau=
  \underbrace{M(q)\ddot q_r+C(q,\dot q)\dot q_r+g(q)-K_Ds}_{\tau_{\mathrm{base}}}
  +\Delta\tau_{\mathrm{TAM}}.
\end{equation*}
The damping gain $K_D\in\mathbb{R}^{N\times N}$ is constant and symmetric positive definite, and $\Delta\tau_{\mathrm{TAM}}(t)\in\mathbb{R}^N$ is TAM's induced additive joint torque.
Substituting this controller into the robot dynamics and using the expressions for $s$ and $\dot s$ gives
\begin{equation}
  M(q)\dot s+C(q,\dot q)s+K_Ds=d_0+\Delta\tau_{\mathrm{TAM}}.
  \label{eq:app_filtered_error_dynamics}
\end{equation}
This passivity-based tracking form is the controller condition used in the proof; a generic PD or OSC controller need not yield Eq.~\eqref{eq:app_filtered_error_dynamics}.
We use the standard Coriolis matrix associated with the same rigid-body model, for which $\dot M-2C$ is skew-symmetric~\citep{spong2020robot}.

\paragraph{Bounds and operating conditions.}
For every admissible initial condition at an initial time $t_0\geq0$, assume that the closed-loop differential equations have a unique solution defined for all $t\geq t_0$ (no finite-time blow-up), and that the following bounds hold uniformly in time and over admissible trajectories:
\begin{equation}
  m_1I_N\preceq M(q)\preceq m_2I_N,\qquad
  \lVert d_0\rVert\leq\bar d_0,\qquad
  \lVert\Delta\tau_{\mathrm{TAM}}\rVert\leq\bar\tau_\Delta.
  \label{eq:app_assumption_bounds}
\end{equation}
Here $0<m_1\leq m_2$ are scalar inertia-eigenvalue bounds and $\bar d_0,\bar\tau_\Delta\in\mathbb{R}_{\geq0}$ are scalar torque-norm bounds.
The identity is $I_N\in\mathbb{R}^{N\times N}$; $A\preceq B$ means that $B-A$ is positive semidefinite, and $\lVert\cdot\rVert$ denotes the Euclidean vector norm or its induced matrix norm.

The TAM torque bound in the additive model above can be enforced by clipping TAM's joint-torque output before adding it to $\tau_{\mathrm{base}}$.
For torque-command outputs, the clipping limits must be translated through the actuator model to justify the bound on the induced joint torque.
Final command saturation must be inactive, or its discrepancy must also be included in $d_0$ with a separately justified uniform bound; command clipping alone does not establish these assumptions.

Define the combined driving torque $w(t)\in\mathbb{R}^N$ and its scalar bound $b$ by
\begin{equation}
  w=d_0+\Delta\tau_{\mathrm{TAM}},\qquad
  \lVert w\rVert\leq\bar d_0+\bar\tau_\Delta=:b.
  \label{eq:app_total_torque_bound}
\end{equation}
Let $\lambda_D=\lambda_{\min}(K_D)>0$ and $\lambda_\Lambda=\lambda_{\min}(\Lambda)>0$, where $\lambda_{\min}$ denotes the smallest eigenvalue.
They quantify the weakest damping and filter-gain directions, respectively.

\begin{tamproposition}[Conditional boundedness of tracking errors]
\label{prop:app_boundedness}
\leavevmode\par\noindent
Under the stated assumptions, define the scalar bounds
\begin{equation}
  B_s=\sqrt{\frac{m_2}{m_1}}
  \max\left\{\lVert s(t_0)\rVert,\frac{b}{\lambda_D}\right\},
  \qquad
  B_e=\max\left\{\lVert e(t_0)\rVert,\frac{B_s}{\lambda_\Lambda}\right\}.
  \label{eq:app_tracking_bound_constants}
\end{equation}
For every $t\geq t_0$, the filtered, position, and velocity errors satisfy
\begin{equation}
  \lVert s(t)\rVert\leq B_s,\qquad
  \lVert e(t)\rVert\leq B_e,\qquad
  \lVert\dot e(t)\rVert\leq B_s+\lVert\Lambda\rVert B_e.
  \label{eq:app_tracking_bounds}
\end{equation}
Thus, all three tracking errors remain bounded for all future times.
\end{tamproposition}

\noindent\textit{Proof.}
We first establish an energy ceiling for $s$, then apply the same reasoning to the position-error equation.

\paragraph{Step 1: Damping dominates at large error.}
Choose the scalar Lyapunov function $V:\mathbb{R}^N\times\mathbb{R}^N\to\mathbb{R}_{\geq0}$ as
\begin{equation}
  V(s,q)=\tfrac12s^\top M(q)s,\qquad
  \tfrac12m_1\lVert s\rVert^2\leq V\leq\tfrac12m_2\lVert s\rVert^2.
  \label{eq:app_lyapunov}
\end{equation}
This has the form of kinetic energy, with $s$ replacing joint velocity.
The inertia bounds ensure that a ceiling on this energy also bounds the ordinary error norm.
Differentiating along the closed-loop solution and substituting Eq.~\eqref{eq:app_filtered_error_dynamics} gives
\begin{align}
  \dot V
  &=s^\top M\dot s+\tfrac12s^\top\dot M s \nonumber\\
  &=-s^\top K_Ds+s^\top w+\tfrac12s^\top(\dot M-2C)s \nonumber\\
  &=-s^\top K_Ds+s^\top w.
  \label{eq:app_energy_balance}
\end{align}
The last equality follows because $x^\top Ax=0$ for every skew-symmetric matrix $A$.
Thus, the inertia-variation and Coriolis terms cancel.
Only damping and the driving torque contribute to the energy derivative.

The damping removes energy at least as fast as $\lambda_D\lVert s\rVert^2$, while Cauchy--Schwarz bounds the energy supplied by the torque as $s^\top w\leq b\lVert s\rVert$.
Therefore,
\begin{equation}
  \dot V\leq-\lambda_D\lVert s\rVert^2+b\lVert s\rVert
  =-\lVert s\rVert\bigl(\lambda_D\lVert s\rVert-b\bigr).
  \label{eq:app_energy_inequality}
\end{equation}
The quadratic damping term dominates the linear torque term whenever $\lVert s\rVert>b/\lambda_D$.
At such errors, $\dot V<0$: the error energy must decrease.

\paragraph{Step 2: An energy ceiling bounds the filtered error.}
Define the scalar energy threshold
\begin{equation}
  V_*=\tfrac12m_2\left(\frac{b}{\lambda_D}\right)^2.
  \label{eq:app_energy_threshold}
\end{equation}
If $V>V_*$, the upper inertia bound in Eq.~\eqref{eq:app_lyapunov} implies $\lVert s\rVert>b/\lambda_D$, and hence $\dot V<0$.
The energy cannot rise above both its initial value and $V_*$: any excursion above that level would require an increase while its derivative is strictly negative.
Consequently,
\begin{equation}
  V(t)\leq\max\{V(t_0),V_*\},
  \qquad t\geq t_0.
  \label{eq:app_energy_ceiling}
\end{equation}
Using the lower inertia bound at time $t$ and the upper inertia bound at $t_0$ gives
\begin{align}
  \lVert s(t)\rVert^2
  &\leq\frac{2}{m_1}\max\{V(t_0),V_*\} \nonumber\\
  &\leq\frac{m_2}{m_1}
  \max\left\{\lVert s(t_0)\rVert^2,
  \left(\frac{b}{\lambda_D}\right)^2\right\}
  =B_s^2.
  \label{eq:app_filtered_error_bound}
\end{align}
This proves $\lVert s(t)\rVert\leq B_s$ for all future times.

\paragraph{Step 3: Bound position and velocity errors.}
To bound the position error, define $E(e)=\tfrac12\lVert e\rVert^2$.
Since $\lVert s(t)\rVert\leq B_s$,
\begin{equation}
  \dot E=e^\top\dot e
  =-e^\top\Lambda e+e^\top s
  \leq-\lambda_\Lambda\lVert e\rVert^2+B_s\lVert e\rVert.
  \label{eq:app_position_energy_inequality}
\end{equation}
Whenever $\lVert e\rVert>B_s/\lambda_\Lambda$, $E$ decreases.
Thus, the position-error magnitude cannot grow beyond the larger of its initial magnitude and this threshold.
Therefore,
\begin{equation}
  \lVert e(t)\rVert\leq
  \max\left\{\lVert e(t_0)\rVert,\frac{B_s}{\lambda_\Lambda}\right\}
  =B_e.
  \label{eq:app_position_bound}
\end{equation}
Finally, $\dot e=s-\Lambda e$ and the triangle inequality give
\begin{equation}
  \lVert\dot e(t)\rVert\leq
  \lVert s(t)\rVert+\lVert\Lambda\rVert\lVert e(t)\rVert
  \leq B_s+\lVert\Lambda\rVert B_e.
  \label{eq:app_velocity_bound}
\end{equation}
These estimates prove Eq.~\eqref{eq:app_tracking_bounds}.
\hfill$\square$

\paragraph{Interpretation for TAM.}
The controller's damping sets an energy ceiling above which any bounded torque input is dominated.
This ceiling keeps filtered, position, and velocity tracking errors finite under the stated assumptions.
The bounds include the initial tracking errors; no convergence rate or settling time is asserted.
For fixed initial errors and $\bar d_0$, increasing the permitted TAM torque can enlarge the conservative bound, while increasing the damping margin can reduce it.
The proof uses only boundedness of TAM's induced torque and does not require accurate predictions at every instant.
When TAM compensates model mismatch, the actual combined torque $d_0+\Delta\tau_{\mathrm{TAM}}$ can be smaller than the sum of their separate bounds, allowing smaller tracking errors than this worst-case estimate.
The proposition concerns conditional tracking-error boundedness and does not certify collision avoidance, contact-force limits, or state constraints.

\subsection{Empirical Oscillation Analysis on the Real Panda}
\label{app:oscillation_spectrum}

As an empirical check for TAM-induced oscillations, we compare the
frequency content of the joint-position tracking error with and without TAM
on the real Panda.
We execute ten paired rollouts of the same reference trajectories with and
without TAM and compute the amplitude spectrum of the joint-position tracking
error, aggregated as RMS per $1$ Hz band over the measured $0$--$200$ Hz
range.
Figure~\ref{fig:app_oscillation_spectrum} shows the result.
The TAM spectrum exhibits no dominant joint-position-error peak anywhere in
the measured band, including at the $5$ Hz latent-update rate of the history
encoder.
This supports the absence of a dominant oscillation peak in these tested trajectories, but does not exclude oscillations under other operating conditions.

\begin{figure}[!htb]
  \centering
  \includegraphics[width=0.85\linewidth]{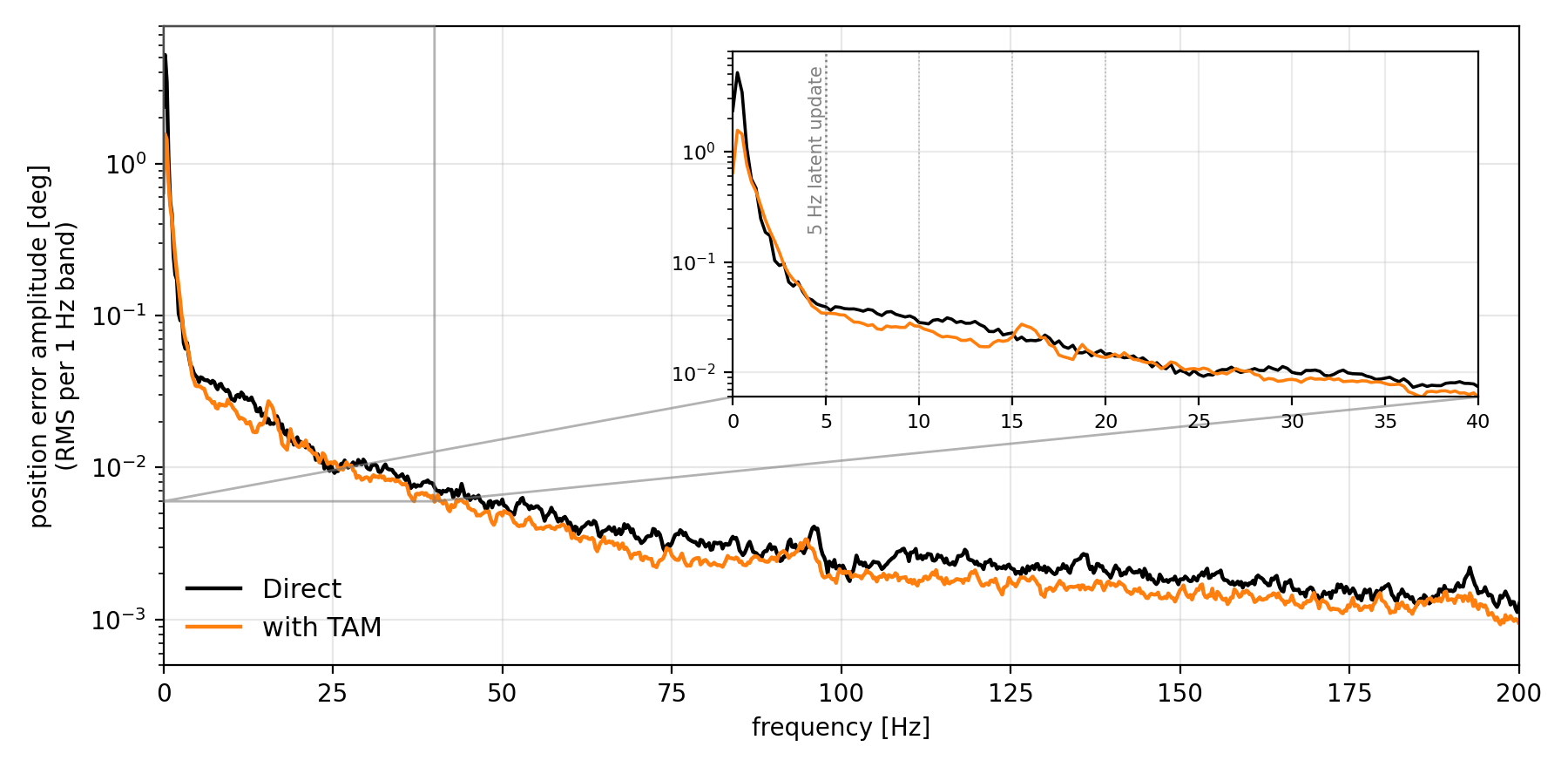}
  \caption{Amplitude spectrum of the joint-position tracking error on the
  real Panda, aggregated as RMS per $1$ Hz band over ten paired rollouts of
  the same reference trajectories executed with and without TAM. The inset
  magnifies the low-frequency band around the $5$ Hz latent-update rate
  (dotted line). No dominant joint-position-error peak appears over
  $0$--$200$ Hz, including at the $5$ Hz latent-update rate.}
  \label{fig:app_oscillation_spectrum}
\end{figure}

\FloatBarrier

\FloatBarrier
\finishmaincompact
\else
\FloatBarrier
\finishmaincompact

\bibliography{ref}  %

\begin{thebibliography}{46}
\providecommand{\natexlab}[1]{#1}
\providecommand{\url}[1]{\texttt{#1}}
\expandafter\ifx\csname urlstyle\endcsname\relax
  \providecommand{\doi}[1]{doi: #1}\else
  \providecommand{\doi}{doi: \begingroup \urlstyle{rm}\Url}\fi

\bibitem[Xie et~al.(2021)Xie, Da, van~de Panne, Babich, and
  Garg]{xie2021dynamics}
Z.~Xie, X.~Da, M.~van~de Panne, B.~Babich, and A.~Garg.
\newblock Dynamics randomization revisited: A case study for quadrupedal
  locomotion.
\newblock In \emph{Proceedings of the IEEE International Conference on Robotics
  and Automation (ICRA)}, pages 4955--4961, 2021.
\newblock \doi{10.1109/ICRA48506.2021.9560837}.

\bibitem[Gaz et~al.(2019)Gaz, Cognetti, Oliva, Giordano, and
  De~Luca]{gaz2019dynamic}
C.~Gaz, M.~Cognetti, A.~Oliva, P.~R. Giordano, and A.~De~Luca.
\newblock Dynamic identification of the {Franka Emika Panda} robot with
  retrieval of feasible parameters using penalty-based optimization.
\newblock \emph{IEEE Robotics and Automation Letters}, 4\penalty0 (4):\penalty0
  4147--4154, 2019.
\newblock \doi{10.1109/LRA.2019.2931248}.

\bibitem[Peng et~al.(2018)Peng, Andrychowicz, Zaremba, and
  Abbeel]{peng2018simtoreal}
X.~B. Peng, M.~Andrychowicz, W.~Zaremba, and P.~Abbeel.
\newblock Sim-to-real transfer of robotic control with dynamics randomization.
\newblock In \emph{Proceedings of the IEEE International Conference on Robotics
  and Automation (ICRA)}, pages 1--8, 2018.
\newblock \doi{10.1109/ICRA.2018.8460528}.

\bibitem[Stone et~al.(1986)Stone, Sanderson, and Neuman]{stone1986arm}
H.~W. Stone, A.~C. Sanderson, and C.~P. Neuman.
\newblock Arm signature identification.
\newblock In \emph{Proceedings of the IEEE International Conference on Robotics
  and Automation (ICRA)}, pages 41--48, April 1986.

\bibitem[Farsoni et~al.(2018)Farsoni, Landi, Ferraguti, Secchi, and
  Bonf{\`e}]{farsoni2018realtime}
S.~Farsoni, C.~T. Landi, F.~Ferraguti, C.~Secchi, and M.~Bonf{\`e}.
\newblock Real-time identification of robot payload using a multirate
  quaternion-based kalman filter and recursive total least-squares.
\newblock In \emph{Proceedings of the IEEE International Conference on Robotics
  and Automation (ICRA)}, pages 2103--2109, 2018.
\newblock \doi{10.1109/ICRA.2018.8461167}.

\bibitem[Swevers et~al.(1997)Swevers, Ganseman, Tukel, De~Schutter, and
  Van~Brussel]{swevers1997optimal}
J.~Swevers, C.~Ganseman, D.~B. Tukel, J.~De~Schutter, and H.~Van~Brussel.
\newblock Optimal robot excitation and identification.
\newblock \emph{IEEE Transactions on Robotics and Automation}, 13\penalty0
  (5):\penalty0 730--740, 1997.
\newblock \doi{10.1109/70.631234}.

\bibitem[Hwangbo et~al.(2019)Hwangbo, Lee, Dosovitskiy, Bellicoso, Tsounis,
  Koltun, and Hutter]{hwangbo2019learning}
J.~Hwangbo, J.~Lee, A.~Dosovitskiy, D.~Bellicoso, V.~Tsounis, V.~Koltun, and
  M.~Hutter.
\newblock Learning agile and dynamic motor skills for legged robots.
\newblock \emph{Science Robotics}, 4\penalty0 (26):\penalty0 eaau5872, 2019.
\newblock \doi{10.1126/scirobotics.aau5872}.

\bibitem[Zhang et~al.(2025)Zhang, Liu, Huang, Han, Lyu, Xu, and
  Zhao]{zhang2025dynamicsprompts}
X.~Zhang, S.~Liu, P.~Huang, W.~J. Han, Y.~Lyu, M.~Xu, and D.~Zhao.
\newblock Dynamics as prompts: In-context learning for sim-to-real system
  identifications.
\newblock In \emph{MARW at AAAI 2025}, 2025.
\newblock URL \url{https://openreview.net/forum?id=8A4lYAGe4u}.

\bibitem[Ramos et~al.(2019)Ramos, Possas, and Fox]{ramos2019bayessim}
F.~Ramos, R.~Possas, and D.~Fox.
\newblock {BayesSim}: Adaptive domain randomization via probabilistic inference
  for robotics simulators.
\newblock In \emph{Proceedings of Robotics: Science and Systems}, Freiburg im
  Breisgau, Germany, June 2019.
\newblock \doi{10.15607/RSS.2019.XV.029}.

\bibitem[{OpenAI} et~al.(2020){OpenAI}, Andrychowicz, Baker, Chociej,
  J{\'o}zefowicz, McGrew, Pachocki, Petron, Plappert, Powell, Ray, Schneider,
  Sidor, Tobin, Welinder, Weng, and Zaremba]{andrychowicz2020learning}
{OpenAI}, M.~Andrychowicz, B.~Baker, M.~Chociej, R.~J{\'o}zefowicz, B.~McGrew,
  J.~Pachocki, A.~Petron, M.~Plappert, G.~Powell, A.~Ray, J.~Schneider,
  S.~Sidor, J.~Tobin, P.~Welinder, L.~Weng, and W.~Zaremba.
\newblock Learning dexterous in-hand manipulation.
\newblock \emph{The International Journal of Robotics Research}, 39\penalty0
  (1):\penalty0 3--20, 2020.
\newblock \doi{10.1177/0278364919887447}.

\bibitem[Muratore et~al.(2022)Muratore, Ramos, Turk, Yu, Gienger, and
  Peters]{muratore2022robot}
F.~Muratore, F.~Ramos, G.~Turk, W.~Yu, M.~Gienger, and J.~Peters.
\newblock Robot learning from randomized simulations: A review.
\newblock \emph{Frontiers in Robotics and AI}, 9:\penalty0 799893, 2022.
\newblock \doi{10.3389/frobt.2022.799893}.

\bibitem[Yu et~al.(2017)Yu, Tan, Liu, and Turk]{yu2017preparing}
W.~Yu, J.~Tan, C.~K. Liu, and G.~Turk.
\newblock Preparing for the unknown: Learning a universal policy with online
  system identification.
\newblock In \emph{Proceedings of Robotics: Science and Systems}, Cambridge,
  Massachusetts, July 2017.
\newblock \doi{10.15607/RSS.2017.XIII.048}.

\bibitem[Kumar et~al.(2021)Kumar, Fu, Pathak, and Malik]{kumar2021rma}
A.~Kumar, Z.~Fu, D.~Pathak, and J.~Malik.
\newblock {RMA}: Rapid motor adaptation for legged robots.
\newblock In \emph{Proceedings of Robotics: Science and Systems}, Virtual, July
  2021.
\newblock \doi{10.15607/RSS.2021.XVII.011}.

\bibitem[Fu et~al.(2022)Fu, Cheng, and Pathak]{fu2022deepwholebody}
Z.~Fu, X.~Cheng, and D.~Pathak.
\newblock Deep whole-body control: Learning a unified policy for manipulation
  and locomotion.
\newblock In \emph{Proceedings of the Conference on Robot Learning (CoRL)},
  2022.
\newblock \doi{10.48550/arXiv.2210.10044}.
\newblock URL \url{https://arxiv.org/abs/2210.10044}.

\bibitem[Liang et~al.(2024)Liang, Ellis, and Henriques]{liang2024rapid}
Y.~Liang, K.~Ellis, and J.~F. Henriques.
\newblock Rapid motor adaptation for robotic manipulator arms.
\newblock In \emph{Proceedings of the IEEE/CVF Conference on Computer Vision
  and Pattern Recognition (CVPR)}, pages 16404--16413, June 2024.
\newblock \doi{10.1109/CVPR52733.2024.01552}.

\bibitem[Vaswani et~al.(2017)Vaswani, Shazeer, Parmar, Uszkoreit, Jones, Gomez,
  Kaiser, and Polosukhin]{vaswani2017attention}
A.~Vaswani, N.~Shazeer, N.~Parmar, J.~Uszkoreit, L.~Jones, A.~N. Gomez,
  L.~Kaiser, and I.~Polosukhin.
\newblock Attention is all you need.
\newblock In \emph{Advances in Neural Information Processing Systems},
  volume~30, pages 5998--6008, 2017.

\bibitem[Nie et~al.(2023)Nie, Nguyen, Sinthong, and Kalagnanam]{nie2023time}
Y.~Nie, N.~H. Nguyen, P.~Sinthong, and J.~Kalagnanam.
\newblock A time series is worth 64 words: Long-term forecasting with
  transformers.
\newblock In \emph{International Conference on Learning Representations
  (ICLR)}, 2023.
\newblock URL \url{https://openreview.net/forum?id=Jbdc0vTOcol}.

\bibitem[Perez et~al.(2018)Perez, Strub, de~Vries, Dumoulin, and
  Courville]{perez2018film}
E.~Perez, F.~Strub, H.~de~Vries, V.~Dumoulin, and A.~Courville.
\newblock {FiLM}: Visual reasoning with a general conditioning layer.
\newblock In \emph{Proceedings of the AAAI Conference on Artificial
  Intelligence}, 2018.
\newblock URL \url{https://arxiv.org/abs/1709.07871}.

\bibitem[Peebles and Xie(2023)]{peebles2023dit}
W.~Peebles and S.~Xie.
\newblock Scalable diffusion models with transformers.
\newblock In \emph{Proceedings of the IEEE/CVF International Conference on
  Computer Vision}, pages 4195--4205, 2023.
\newblock \doi{10.1109/ICCV51070.2023.00387}.

\bibitem[Ross et~al.(2011)Ross, Gordon, and Bagnell]{ross2011reduction}
S.~Ross, G.~Gordon, and D.~Bagnell.
\newblock A reduction of imitation learning and structured prediction to
  no-regret online learning.
\newblock In \emph{Proceedings of the Fourteenth International Conference on
  Artificial Intelligence and Statistics}, volume~15 of \emph{Proceedings of
  Machine Learning Research}, pages 627--635, 2011.
\newblock URL \url{https://proceedings.mlr.press/v15/ross11a.html}.

\bibitem[Zakka et~al.(2025)Zakka, Tabanpour, Liao, Haiderbhai, Holt, Luo,
  Allshire, Frey, Sreenath, Kahrs, Sferrazza, Tassa, and
  Abbeel]{zakka2025mujocoplayground}
K.~Zakka, B.~Tabanpour, Q.~Liao, M.~Haiderbhai, S.~Holt, J.~Y. Luo,
  A.~Allshire, E.~Frey, K.~Sreenath, L.~A. Kahrs, C.~Sferrazza, Y.~Tassa, and
  P.~Abbeel.
\newblock {MuJoCo Playground}.
\newblock In \emph{Robotics: Science and Systems (RSS)}, 2025.
\newblock URL \url{https://playground.mujoco.org/}.

\bibitem[Zhao et~al.(2023)Zhao, Kumar, Levine, and Finn]{zhao2023act}
T.~Z. Zhao, V.~Kumar, S.~Levine, and C.~Finn.
\newblock Learning fine-grained bimanual manipulation with low-cost hardware.
\newblock In \emph{Proceedings of Robotics: Science and Systems}, 2023.
\newblock \doi{10.15607/RSS.2023.XIX.016}.

\bibitem[Howell et~al.(2022)Howell, Gileadi, Tunyasuvunakool, Zakka, Erez, and
  Tassa]{howell2022predictive}
T.~Howell, N.~Gileadi, S.~Tunyasuvunakool, K.~Zakka, T.~Erez, and Y.~Tassa.
\newblock Predictive sampling: Real-time behaviour synthesis with {MuJoCo}.
\newblock \emph{arXiv preprint arXiv:2212.00541}, 2022.
\newblock \doi{10.48550/arXiv.2212.00541}.
\newblock URL \url{https://arxiv.org/abs/2212.00541}.

\bibitem[Zakka et~al.(2022)Zakka, Tassa, and {MuJoCo Menagerie
  Contributors}]{zakka2022mujocomenagerie}
K.~Zakka, Y.~Tassa, and {MuJoCo Menagerie Contributors}.
\newblock {MuJoCo Menagerie}: A collection of high-quality simulation models
  for {MuJoCo}, 2022.
\newblock URL \url{https://github.com/google-deepmind/mujoco_menagerie}.

\bibitem[Memmel et~al.(2024)Memmel, Wagenmaker, Zhu, Yin, Fox, and
  Gupta]{memmel2024asid}
M.~Memmel, A.~Wagenmaker, C.~Zhu, P.~Yin, D.~Fox, and A.~Gupta.
\newblock {ASID}: Active exploration for system identification in robotic
  manipulation.
\newblock In \emph{Proceedings of the International Conference on Learning
  Representations (ICLR)}, 2024.
\newblock URL \url{https://arxiv.org/abs/2404.12308}.

\bibitem[Fey et~al.(2025)Fey, Margolis, Peticco, and Agrawal]{fey2025uan}
N.~Fey, G.~B. Margolis, M.~Peticco, and P.~Agrawal.
\newblock Bridging the sim-to-real gap for athletic loco-manipulation.
\newblock \emph{arXiv preprint arXiv:2502.10894}, 2025.
\newblock \doi{10.48550/arXiv.2502.10894}.
\newblock URL \url{https://arxiv.org/abs/2502.10894}.

\bibitem[Liu et~al.(2025)Liu, Wang, and Yi]{liu2025dexndm}
X.~Liu, H.~Wang, and L.~Yi.
\newblock {DexNDM}: Closing the reality gap for dexterous in-hand rotation via
  joint-wise neural dynamics model.
\newblock \emph{arXiv preprint arXiv:2510.08556}, 2025.
\newblock \doi{10.48550/arXiv.2510.08556}.

\bibitem[{Lennart Ljung}(1999)]{ljung1999system}
{Lennart Ljung}.
\newblock {System Identification}, 1999.

\bibitem[Qi et~al.(2022)Qi, Kumar, Calandra, Ma, and Malik]{qi2022inhand}
H.~Qi, A.~Kumar, R.~Calandra, Y.~Ma, and J.~Malik.
\newblock In-hand object rotation via rapid motor adaptation.
\newblock In \emph{Proceedings of the Conference on Robot Learning (CoRL)},
  pages 1722--1732, 2022.
\newblock URL \url{https://proceedings.mlr.press/v205/qi23a.html}.

\bibitem[Liu et~al.(2025)Liu, Pathak, and Agarwal]{liu2025locoformer}
M.~Liu, D.~Pathak, and A.~Agarwal.
\newblock {LocoFormer}: Generalist locomotion via long-context adaptation.
\newblock \emph{arXiv preprint arXiv:2509.23745}, 2025.
\newblock \doi{10.48550/arXiv.2509.23745}.

\bibitem[Zhang et~al.(2025)Zhang, Guo, Chen, Wang, Lin, Lian, Xue, Wang, Liu,
  Liu, Wang, and Yi]{zhang2025any2track}
Z.~Zhang, J.~Guo, C.~Chen, J.~Wang, C.~Lin, Y.~Lian, H.~Xue, Z.~Wang, M.~Liu,
  H.~Liu, H.~Wang, and L.~Yi.
\newblock Track any motions under any disturbances.
\newblock \emph{arXiv preprint arXiv:2509.13833}, 2025.
\newblock \doi{10.48550/arXiv.2509.13833}.

\bibitem[Slotine and Li(1987)]{slotine1987adaptive}
J.-J.~E. Slotine and W.~Li.
\newblock On the adaptive control of robot manipulators.
\newblock \emph{The International Journal of Robotics Research}, 6\penalty0
  (3):\penalty0 49--59, 1987.
\newblock \doi{10.1177/027836498700600303}.

\bibitem[Lyu et~al.(2024)Lyu, Lang, Zhao, Zhang, Ding, and Wang]{lyu2024rl2ac}
S.~Lyu, X.~Lang, H.~Zhao, H.~Zhang, P.~Ding, and D.~Wang.
\newblock {RL2AC}: Reinforcement learning-based rapid online adaptive control
  for legged robot robust locomotion.
\newblock In \emph{Proceedings of Robotics: Science and Systems}, 2024.
\newblock \doi{10.15607/RSS.2024.XX.060}.

\bibitem[Joukov et~al.(2015)Joukov, Bonnet, Venture, and
  Kuli{\'c}]{joukov2015constrained}
V.~Joukov, V.~Bonnet, G.~Venture, and D.~Kuli{\'c}.
\newblock Constrained dynamic parameter estimation using the extended kalman
  filter.
\newblock In \emph{Proceedings of the IEEE/RSJ International Conference on
  Intelligent Robots and Systems (IROS)}, pages 3654--3659, 2015.
\newblock \doi{10.1109/IROS.2015.7353888}.

\bibitem[Hanna and Stone(2017)]{hanna2017grounded}
J.~Hanna and P.~Stone.
\newblock Grounded action transformation for robot learning in simulation.
\newblock In \emph{Proceedings of the 31st AAAI Conference on Artificial
  Intelligence (AAAI)}, San Francisco, CA, February 2017.

\bibitem[Hanna et~al.(2021)Hanna, Desai, Karnan, Warnell, and
  Stone]{hanna2021grounded}
J.~P. Hanna, S.~Desai, H.~Karnan, G.~Warnell, and P.~Stone.
\newblock Grounded action transformation for sim-to-real reinforcement
  learning.
\newblock \emph{Machine Learning}, 110:\penalty0 2469--2499, 2021.
\newblock \doi{10.1007/s10994-021-05982-z}.

\bibitem[Karnan et~al.(2020)Karnan, Desai, Hanna, Warnell, and
  Stone]{karnan2020reinforced}
H.~Karnan, S.~Desai, J.~P. Hanna, G.~Warnell, and P.~Stone.
\newblock Reinforced grounded action transformation for sim-to-real transfer.
\newblock In \emph{Proceedings of the IEEE/RSJ International Conference on
  Intelligent Robots and Systems (IROS)}, 2020.
\newblock URL \url{https://arxiv.org/abs/2008.01279}.

\bibitem[Desai et~al.(2020)Desai, Karnan, Hanna, Warnell, and
  Stone]{desai2020stochastic}
S.~Desai, H.~Karnan, J.~P. Hanna, G.~Warnell, and P.~Stone.
\newblock Stochastic grounded action transformation for robot learning in
  simulation.
\newblock In \emph{Proceedings of the IEEE/RSJ International Conference on
  Intelligent Robots and Systems (IROS)}, 2020.
\newblock URL \url{https://arxiv.org/abs/2008.01281}.

\bibitem[Johannink et~al.(2019)Johannink, Bahl, Nair, Luo, Kumar, Loskyll,
  Ojea, Solowjow, and Levine]{johannink2019residual}
T.~Johannink, S.~Bahl, A.~Nair, J.~Luo, A.~Kumar, M.~Loskyll, J.~A. Ojea,
  E.~Solowjow, and S.~Levine.
\newblock Residual reinforcement learning for robot control.
\newblock In \emph{Proceedings of the IEEE International Conference on Robotics
  and Automation (ICRA)}, pages 6023--6029, 2019.
\newblock \doi{10.1109/ICRA.2019.8794127}.

\bibitem[He et~al.(2025)He, Gao, Xiao, Zhang, Wang, Wang, Luo, He, Sobanbabu,
  Pan, Yi, Qu, Kitani, Hodgins, Fan, Zhu, Liu, and Shi]{he2025asap}
T.~He, J.~Gao, W.~Xiao, Y.~Zhang, Z.~Wang, J.~Wang, Z.~Luo, G.~He,
  N.~Sobanbabu, C.~Pan, Z.~Yi, G.~Qu, K.~Kitani, J.~Hodgins, L.~J. Fan, Y.~Zhu,
  C.~Liu, and G.~Shi.
\newblock {ASAP}: Aligning simulation and real-world physics for learning agile
  humanoid whole-body skills.
\newblock \emph{arXiv preprint arXiv:2502.01143}, 2025.

\bibitem[Jiang et~al.(2024)Jiang, Wang, Zhang, Wu, and
  Fei-Fei]{jiang2024transic}
Y.~Jiang, C.~Wang, R.~Zhang, J.~Wu, and L.~Fei-Fei.
\newblock {TRANSIC}: Sim-to-real policy transfer by learning from online
  correction.
\newblock In \emph{Proceedings of the Conference on Robot Learning (CoRL)},
  2024.
\newblock URL \url{https://arxiv.org/abs/2405.10315}.

\bibitem[Todorov et~al.(2012)Todorov, Erez, and Tassa]{todorov2012mujoco}
E.~Todorov, T.~Erez, and Y.~Tassa.
\newblock {MuJoCo}: A physics engine for model-based control.
\newblock In \emph{Proceedings of the IEEE/RSJ International Conference on
  Intelligent Robots and Systems (IROS)}, pages 5026--5033, 2012.
\newblock \doi{10.1109/IROS.2012.6386109}.

\bibitem[Su et~al.(2021)Su, Lu, Pan, Murtadha, Wen, and Liu]{su2021roformer}
J.~Su, Y.~Lu, S.~Pan, A.~Murtadha, B.~Wen, and Y.~Liu.
\newblock {RoFormer}: Enhanced transformer with rotary position embedding.
\newblock \emph{arXiv preprint arXiv:2104.09864}, 2021.
\newblock URL \url{https://arxiv.org/abs/2104.09864}.

\bibitem[Qi et~al.(2017)Qi, Su, Mo, and Guibas]{qi2017pointnet}
C.~R. Qi, H.~Su, K.~Mo, and L.~J. Guibas.
\newblock {PointNet}: Deep learning on point sets for 3d classification and
  segmentation.
\newblock In \emph{Proceedings of the IEEE Conference on Computer Vision and
  Pattern Recognition (CVPR)}, pages 652--660, 2017.

\bibitem[Thrun et~al.(2005)Thrun, Burgard, and Fox]{thrun2005probabilistic}
S.~Thrun, W.~Burgard, and D.~Fox.
\newblock \emph{Probabilistic Robotics}.
\newblock MIT Press, Cambridge, MA, 2005.
\newblock ISBN 9780262201629.

\bibitem[Spong et~al.(2020)Spong, Hutchinson, and Vidyasagar]{spong2020robot}
M.~W. Spong, S.~Hutchinson, and M.~Vidyasagar.
\newblock \emph{Robot Modeling and Control}.
\newblock Wiley, 2 edition, 2020.

\end{thebibliography}
\fi

\end{document}